\documentclass{article}

\usepackage{PRIMEarxiv}

\usepackage[utf8]{inputenc} 
\usepackage[T1]{fontenc}    
\usepackage{hyperref}       
\usepackage{url}
\usepackage{booktabs}       
\usepackage{amsfonts}       
\usepackage{nicefrac}       
\usepackage{microtype}      
\usepackage{lipsum}
\usepackage{fancyhdr}       
\usepackage{graphicx}       
\usepackage{graphicx,caption,subcaption} 
\usepackage{comment}
\usepackage{amsmath}

\newcommand{\norm}[1]{\left\lVert #1 \right\rVert}
\newcommand{\diag}{\operatorname{diag}}
\newcommand{\vect}[1]{\mathbf{#1}}

\usepackage{cleveref}
\Crefformat{figure}{#2Fig.~#1#3}
\Crefmultiformat{figure}{Figs.~#2#1#3}{ and~#2#1#3}{, #2#1#3}{ and~#2#1#3}
\Crefformat{equation}{#2Eq.~#1#3}
\Crefmultiformat{equation}{Eqs.~#2#1#3}{ and~#2#1#3}{, #2#1#3}{ and~#2#1#3}

\newcommand{\beginsupplement}{%
        \setcounter{table}{0}
        \renewcommand{\thetable}{S\arabic{table}}%
        \setcounter{figure}{0}
        \renewcommand{\thefigure}{S\arabic{figure}}%
        \setcounter{equation}{0}
        \renewcommand{\theequation}{S\arabic{equation}}
        \setcounter{section}{0}
     }

\usepackage{bibunits}
\defaultbibliography{Catoni_et_al} 
\defaultbibliographystyle{unsrt} 
    
\graphicspath{{figures/}}     
\makeatletter
\renewcommand{\fnum@figure}{Fig. \thefigure}
\makeatother
\title{Remedying uncertainty representations in visual inference through Explaining-Away Variational Autoencoders
}

\author{
 Josefina Catoni $^{1.,*}$ \\
  \href{mailto:jcatoni@sinc.unl.edu.ar}{jcatoni@sinc.unl.edu.ar} \\
\And
 Domonkos Martos $^{2.,*}$ \\
  \href{mailto:martos.domonkos@wigner.hun-ren.hu}{martos.domonkos@wigner.hun-ren.hu}
\And
 Ferenc Csikor $^{2.}$ \\
\And
 Enzo Ferrante $^{1.,3.}$ \\
\And
 Diego H. Milone $^{1.}$ \\
\And 
 Balázs Meszéna $^{2.}$  \\
\And 
 Gergő Orbán $^{2.,\dag}$  \\
\And 
 Rodrigo Echeveste $^{1.,\dag}$ \\
}

\begin{document}
\begin{bibunit}
\maketitle

\begin{abstract}
Optimal computations under uncertainty require an adequate probabilistic representation about beliefs. Deep generative models, and specifically Variational Autoencoders (VAEs), have the potential to meet this demand by building latent representations that learn to associate uncertainties with inferences while avoiding their characteristic intractable computations. Yet, we show that it is precisely uncertainty representation that suffers from inconsistencies under an array of relevant computer vision conditions: contrast-dependent computations, image corruption, out-of-distribution detection. Drawing inspiration from classical computer vision, we present a principled extension to the standard VAE by introducing a simple yet powerful inductive bias through a global scaling latent variable, which we call the Explaining-Away VAE (EA-VAE). By applying EA-VAEs to a spectrum of computer vision domains and a variety of datasets, spanning standard NIST datasets to rich medical and natural image sets, we show the EA-VAE restores normative requirements for uncertainty. Furthermore, we provide an analytical underpinning of the contribution of the introduced scaling latent to contrast-related and out-of-distribution related modulations of uncertainty, demonstrating that this mild inductive bias has stark benefits in a broad set of problems. 
Moreover, we find that EA-VAEs recruit divisive normalization, a motif widespread in biological neural networks, to remedy defective inference.
Our results demonstrate that an easily implemented, still powerful update to the VAE architecture can remedy defective inference of uncertainty in probabilistic computations.
\end{abstract}


\section{Introduction}
An adequate representation of uncertainty is key to reliable decision making and sensory integration \cite{MacKay2003}, and thus its importance for artificial systems is increasingly evident as flexibility, adaptation and multimodality become widespread requirements. Maintaining an estimate of the uncertainty associated with inferences permits better reasoning about beliefs \cite{JaynesProbTheory}, generalization \cite{Gawlikowski2023}, and assessment of risks associated with alternative decisions \cite{Berger1980}. Furthermore, it allows for more efficient use of acquired knowledge for learning \cite{gal2016dropout, Osband16Neurips, Houlsby2011}, and downstream computations in a hierarchical processing pipeline, among others. In particular, uncertainty-aware AI systems are argued to be necessary for exploring novel domains \cite{Wang2023, JimenezLuna2020}. Indeed, probabilistic machine learning promises to deliver tools that are not only capable of inferring best estimates but also the associated confidence in those estimates \cite{ghahramani2015}. In this context, generative models have been advocated as a central framework for probabilistic machine learning \cite{bishop2006pattern, Tomczak2022DeepGenMod}. Recently, deep learning has been elevated to accommodate deep generative models, including generative adversarial networks \cite{goodfellow2014GANs}, variational autoencoders (VAEs)  \cite{kingma2022autoencoding,rezende2016}, and diffusion models \cite{KadkhodaieSimoncelli2021, song2021, sohldickstein2015}.
In particular, when learning a latent representation of the data, VAEs provide an explicit representation of the uncertainty of inferences, rendering them a central tool for applications that require probabilistic computations \cite{kingma2022autoencoding, higgins2017betavae, LopezRegierColeJordanYosef2018, vandenOordVinyalsKavukcouglu}. 

\begin{figure*}[t!]
\centering
\includegraphics[width=1.0\linewidth]{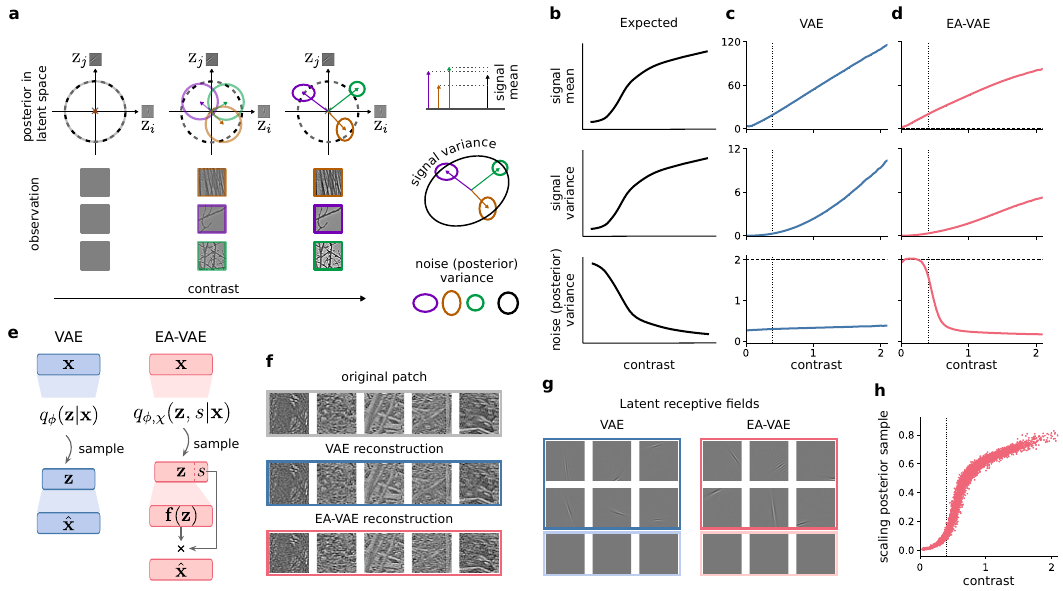}
\caption{\textbf{Characteristic properties of inferred posteriors as contrast is varied.} \textbf{a}, Expected behavior of the posteriors of a pair of latent variables in natural images when contrast is systematically increased. Three example images (\emph{green}, \emph{purple} and \emph{mustard frames}) are presented at varying contrast levels. As contrast increases, the inferred posteriors for each image (\emph{colored solid lines}) progressively deviate both from the prior (\emph{dashed line}) and from one another. There are two different sources of variance in latent representations. The \emph{signal variance} refers to how the posterior mean changes with the input. The \emph{noise variance} (or posterior variance) refers to the remaining uncertainty after having observed a given stimulus. Similarly, we call \emph{signal mean} the average distance between the prior’s mean and the posteriors’ mean. The \emph{noise mean} will be zero in our models, as is usually assumed for VAEs, without loss of generality. \textbf{b}, Cartoon of expected behavior of signal mean, signal variance and noise variance as a function of contrast showing qualitative trends of these quantities. \textbf{c},\textbf{d}, Signal mean, signal variance and noise variance of the inferred latent posteriors in natural image-trained VAE (\textbf{c}) and EA-VAE (\textbf{d}, respectively), as a function of image contrast (Supplementary \Cref{subsec:uncertainty_latent}). Prior mean and variance are shown in \emph{horizontal dashed black lines}, and observation noise is shown in \emph{vertical dotted black lines}. \textbf{e}, Comparison of the inference and reconstruction process in the VAE (\emph{left}) and EA-VAE (\emph{right}). In the VAE a single pool of latent variables $\mathbf{z}$ is inferred, while in the EA-VAE there is an additional global latent variable $s$ which acts multiplicatively on the latents, $\mathbf{z}$. \textbf{f}, Example test image patches and their respective reconstruction through VAE and EA-VAE. \textbf{g}, Receptive filters of example latents in the VAE and EA-VAE. High-saturation boxes surround informative latent units while low saturation boxes surround non-informative ones (Supplementary \Cref{subsec:receptive_fields}). \textbf{h}, Inferred posterior mean of the scaling variable for individual patches (dots) in the EA-VAE model, as a function of the measured contrast of these images.}
\label{fig:intro}
\hrulefill
\end{figure*}

A key element of probabilistic computations is that evidence, the likelihood, is combined with expectations, the prior, to interpret observations, yielding the posterior distribution whose width carries information about uncertainty of inferences. Given the central role of uncertainty in VAEs, it is crucial to evaluate to what extent uncertainty representations fulfill elementary requirements set by probability theory. Indeed, the capacity of deep generative models to faithfully represent uncertainty has been questioned \cite{nalisnick2018deep}. Focusing on visual inference tasks, we first consider contrast, a salient variable of natural image statistics, and identify four fundamental requirements towards principled inference \cite{berkes2011spontaneous}. First, at zero contrast, where the image carries no information, the inferred posterior needs to be equal to the prior (\Cref{fig:intro}a).

Second, with increasing contrast, observations become progressively more informative and the corresponding posteriors should become increasingly distinct from the prior. Accordingly, the average distance between the means of the posteriors and the prior, which we term \emph{signal mean}, should increase (\Cref{fig:intro}a and b, top). Third, more informative observations yield increasing variance in posteriors means (\Cref{fig:intro}a and b, center), usually called \emph{signal variance}. Finally, with increasing contrast observations become more informative, resulting in decreasing uncertainty about the latent generative component, i.e.\ decreasing posterior variance, referred to as \emph{noise variance} (\Cref{fig:intro}a and b, bottom). However, in this work we show that when training a VAE on natural images (\Cref{fig:intro}f), although contrast-dependence of the posterior mean matches these requirements, contrast-dependent changes in posterior variances are at odds with the promise of a faithful representation of uncertainty (\Cref{fig:intro}, cf. b and c). 

We argue that this signature of inaccurate inference is a consequence of the mismatch between the assumptions about the statistics of latent representations and those characteristic of generative factors underlying naturally occurring inputs. In its basic formulation, latent variables in VAEs are assumed to be either independent or linearly correlated. Natural stimuli, however, have been shown to display strong nonlinear dependencies between latents \cite{NIPS1999_6a5dfac4}.

We propose a theoretically well-motivated and powerful update to the standard VAE architecture, the Explaining-Away-VAE (EA-VAE for short). EA-VAE incorporates an inductive bias in the generative model by learning an extra latent variable, which acts as a global multiplicative scaling that is capable of modulating the variance of the data. This extension was inspired by the Gaussian Scale Mixtures (GSM) model \cite{NIPS1999_6a5dfac4}, a simple yet powerful model of natural images for compression and denoising, which has also been successful in interpreting behavioral and neural data in biological perception \cite{schwartz2001natural, orban2016neural, festa2021GSM, echeveste2020cortical}. As recruiting the scaling variable can explain global changes in the input, the introduced scaling variable can replace correlated modulations of latent activations upon changes in image contrast. Hence the term explaining-away in our EA-VAEs.

In this work, we demonstrate that EA-VAEs address a broad set of issues concerning VAEs, beyond producing contrast-dependent posteriors that match the fundamental requirements of probabilistic inference: the reliable representation of the posterior uncertainty contributes to enhanced inference capabilities in a broad set of tasks. First, we show that limited inference capabilities of VAEs are present in canonical computer vision domains such as handwritten character recognition \cite{mnist} and x-ray medical images \cite{chestmnist}, which are consistently remedied by EA-VAEs. Second, we demonstrate that the introduction of the scaling variable in the generative model shapes learned representations such that it promotes disentanglement of latent variables. Third, we demonstrate that EA-VAEs address challenges beyond contrast-dependent inferences, correcting inference for image manipulations that affect uncertainty and addressing a widely-known challenge of VAEs: the detection of out-of-distribution data. Fourth, we show the benefits of a more faithful uncertainty representation through information fusion. Finally, we identify biologically motivated properties of the recognition model of EA-VAEs that complement the inductive bias introduced in the generative model through the scaling latent variable. We demonstrate that improved inference in EA-VAEs is made possible by computations that bear the signatures of a widely known computational motif in biological neural networks:  divisive normalization \cite{carandini2012normalization, goris24natrevneuro}.  The results open up multiple novel avenues of research, showing great promise for machine learning applications where uncertainty estimates are of importance, such as information fusion or anticipation of failure modes.

\section{Methods}
We start by presenting the formalism of standard VAEs, which we subsequently use to introduce the EA-VAE. VAEs learn about the statistics of data through discovering latent variables, $\mathbf{z} \in \mathbb{R}^D$. Attractively, latent variables can be considered as generative factors underlying data. This relationship between latent variables and data is summarized as a generative model $p(\mathbf{x}|\mathbf{z})$, which is complete with a prior distribution over latent features, $p(\mathbf{z})$. Upon observing a new data point, $\mathbf{x}\in \mathbb{R}^M$, an inference is made by calculating the posterior distribution $p(\mathbf{z}|\mathbf{x})$. Obtaining an exact posterior is in general prohibitive and VAEs establish a highly principled approach to obtain an approximate posterior by learning a recognition model, $q(\mathbf{z}|\mathbf{x})$. VAEs learn in a self-supervised manner,  reconstructing the input via the encoder and decoder models, also referred to as the recognition and the generative model. The encoder and the decoder are parametrized in terms of neural networks with parameters $\phi$ and $\psi$, respectively. For any input $\mathbf{x}$, the encoder calculates the parameters of the variational approximation of the posterior, $q_\phi(\mathbf{z}|\mathbf{x})$. After taking a sample $\mathbf{z}$ from $q_\phi(\mathbf{z}|\mathbf{x})$, the autoencoder attempts to reconstruct the input by synthesizing $\hat{\mathbf{x}}$ through the decoder, $p_\psi(\mathbf{x} | \mathbf{z})$ (\Cref{fig:intro}e, left). Depending on the pixel distribution of the data, the likelihood can be a normal or a Bernoulli distribution, with independent pixels in both cases. 

The objective function to learn the parameters of the encoder and decoder is formulated as the Evidence Lower Bound (ELBO)\cite{kingma2019}. This takes the form

\begin{equation}
\mathcal{L}_{VAE}(\mathbf{x},\phi,\psi) = -\mathbb{E}_{q_\phi(\mathbf{z}|\mathbf{x})} \left[ \log p_\psi(\mathbf{x}|\mathbf{z})\right]
 + \beta D_{KL}(q_\phi(\mathbf{z}|\mathbf{x})||p(\mathbf{z})),
\label{eqn:VAEcost}
\end{equation}

\noindent
where the prior, $p(\mathbf{z})$, is usually a standard normal distribution. The first term expresses the reconstruction error, and the second term acts as regularization, penalizing differences between the prior distribution and the approximate posterior. The hyperparameter $\beta$ determines the prior regularization weight, and is linked to the assumed pixel observation noise $\boldsymbol{\sigma}_{obs}^2$ \cite{rybkin2021}.

The expectation in the first term is not explicitly calculated, but a Monte Carlo approximation is taken with a single sample from the variational posterior. With this approximation, the reconstruction error takes either the form of the cross-entropy or quadratic loss between an input and its reconstruction for the Bernoulli and normal likelihoods, respectively.

\subsection{The explaining-away variational autoencoder (EA-VAE)}

In this paper we propose a novel variational inference architecture, specifically designed to improve uncertainty estimates. To that end, the EA-VAE extends the latent space of standard VAEs with a scalar latent variable $s\in \mathbb{R}^+$ that acts multiplicatively in the decoder. That is, the mapping $\mathbf{f}(\mathbf{z})$ in the generative model of a standard VAE is supplemented by $s$ such that the mean of the likelihood in the EA-VAE is obtained as $s\, \mathbf{f}(\mathbf{z})$ (\Cref{fig:intro}e, right). The scaling variable is inferred simultaneously with the other latents, $\mathbf{z}$. The multiplicative contribution of $s$ to the likelihood implements a division of labor between latents: while traditional latents $\mathbf{z}$ encode a variety of details about the input, $s$ accounts for overall variance in the pixel space. Intuitively, the scaling variable can represent global modulations in an image without changes in the content of the image. This division of labor has the consequence that $s$ is directly in charge of modulating the model's uncertainty about the input. Large values of $s$ encourage the network to represent the input with high fidelity and certainty, while small values of $s$ encourage the posterior of $\mathbf{z}$ to collapse to the prior, reporting maximal uncertainty (see mathematical derivation in Supplementary \Cref{subsec:theoretical}).

For EA-VAEs, the variational posterior is a joint distribution: $q_{\phi,\chi}(\mathbf{z},s |\mathbf{x})$. A widespread approach for VAEs is to learn a variational posterior that assumes independence among the latents. Accordingly, the joint variational posterior of EA-VAE was broken down into two components, $q_{\phi}(\mathbf{z}|\mathbf{x})$ and $q_{\chi}(s |\mathbf{x})$, where $\phi$ and $\chi$ denote the parameters of the two components of the encoder. As $s$ is meant to account for the joint modulation of pixels intensities, a positive value is justified. As shown later, the results are robust against the specific choice of the variational posterior of $s$. We note that whenever a VAE was contrasted with an EA-VAE, the total number of latent variables was preserved to ensure a fair comparison.

The extended generative model yields an updated training objective, that is, an updated ELBO:

\begin{equation}
\mathcal{L}_{EA-VAE}(\mathbf{x},\phi,\psi,\chi) = -\mathbb{E}_{q_\phi(\mathbf{z}|\mathbf{x})} \left[ \log p_\psi(\mathbf{x}|\mathbf{z},s)\right]
 + \beta_1 D_{KL}(q_\phi(\mathbf{z}|\mathbf{x})||p(\mathbf{z}))
 +\beta_2 D_{KL}(q_\chi(s|\mathbf{x})||p(s)).
\label{eqn:EA-VAEcost}
\end{equation}

\noindent Here $\beta_1$ and $\beta_2$ denote the relative strengths of the prior regularization terms of $\mathbf{z}$ and $s$ and respectively.
For reconstruction, both $\mathbf{z}$ and $s$ are sampled from their respective posteriors before decoding.

Next, we present the specific VAE and EA-VAE architectures investigated in the paper.

\subsection{VAEs for natural images}\label{subsec:method-vae-eavae-nat}
We explored the contribution of the scaling variable to inference in VAEs using an architecture that featured a linear generative model. In this model the scaling variable, $s$, can be studied in detail as it has a well-defined role for the representation of contrast. Our analysis is aligned with earlier accounts by adopting a large-dimensional latent space, fully connected layers, and a Laplace prior \cite{geadah2018sparse, csikor2025top}. We trained the model on gray-scale natural image patches.

The Laplace prior serves a sparse coding approach, which has historically been taken to capture natural image statistics and recover the receptive field properties of neurons in the primary visual cortex \cite{Olshausen1996}. We hence employ these models as a starting point for our analysis (though we will thoroughly analyze VAEs with Gaussian priors in later sections). To ensure that the KL divergence in the ELBO is analytically tractable, we employed the Laplace distribution for the variational posterior too: $q_\phi(\mathbf{z}|\mathbf{x}) = \textrm{Laplace}(\mathbf{z};\boldsymbol{\mu}{(\mathbf{x})},\boldsymbol{b}{(\mathbf{x})})$, with $\boldsymbol{\mu}{(\mathbf{x})}$ and $\boldsymbol{b}{(\mathbf{x})}$ being the respective mean and scale parameters. The generative model for a given $\mathbf{x}$ is parameterized by a multivariate uncorrelated Normal distribution $p_\psi(\mathbf{x}|\mathbf{z}) = \mathcal{N}(\mathbf{x};\mathbf{f}(\mathbf{z}),\boldsymbol{\sigma}_{\text{obs}}^2)$,
such that $\mathbf{f}(\mathbf{z}) = \mathbf{A} \cdot \mathbf{z}$ is a linear transformation, and $\sigma_{\text{obs}}$ is the observation noise (see Supplementary \Cref{subsec:implementation} for detailed cost function)

The EA-VAE extension preserves the previous model structure and incorporates the scalar multiplicative latent variable $s$ (cf. \Cref{fig:archs} a and b). To avoid degeneracy in the image reconstruction we assume $s$ to be positive. We implemented a range of variants for this: a softplus activated Laplace distribution, a Log-normal distribution, and a Gamma distribution (see Supplementary \Cref{subsec:implementation} for detailed cost functions). We used the first variant as a default and used the variants as controls.

\subsection{VAEs for other classic computer vision domains}\label{subsec:method-vaeeavae-nist}
We studied the properties of EA-VAEs on standard computer vision datasets (MNIST, ChestMNIST, FashionMNIST, NIH-ChestXray14, Imagenet, see details in \Cref{subsec:datasets}). The corresponding variational autoencoders were not restricted to linear decoders and, as is common practice, convolutional layers were also considered. As is standard in the field, the variational posterior was a multivariate uncorrelated Normal distribution $q_\phi(\mathbf{z}|\mathbf{x}) = \textrm{Normal}(\mathbf{z};\boldsymbol{\mu}{(\mathbf{x})},\boldsymbol{\sigma}^2 (\mathbf{x}))$, with $\boldsymbol{\mu}{(\mathbf{x})}$ and $\boldsymbol{\sigma}^2{(\mathbf{x})}$ the respective mean and variance vectors. In the case of the EA-VAE the variational posterior of the multiplicative variable, $s$, was log-normal: $q_\chi(s|\mathbf{x}) = \textrm{Log-normal}(s;\nu{(\mathbf{x})},\xi^2{(\mathbf{x})})$. Priors $p(\mathbf{z})$ and $p(s)$ come from the same family as their respective variational posteriors but with zero mean and diagonal unit variance.

When studying the models on the MNIST dataset, we assumed a Bernoulli distribution in the decoder. This choice is justified by the largely black-and-white distribution of pixel intensities of digits. For other datasets (see details in  \Cref{subsec:datasets}), the decoder was parameterized by a multivariate uncorrelated Normal distribution (Supplementary \Cref{subsec:implementation} for detailed cost functions).

\section{Experimental Setup}

\subsection{Datasets}\label{subsec:datasets}
\subsubsection{Natural Images}
Following \cite{csikor2022topdown}, we used the Van Hateren dataset, composed of natural image patches of $40\times40$ pixels each \cite{Hateren1998data} ($6.4\times10^5$ images for training and validation and $6.4\times10^4$ for test). The first set was further divided into 80$\%$ for training and 20$\%$ for validation purposes. Image patches had already been preprocessed, whereby $100(1 - \frac{\pi}{4})\%$ of the higher frequency PCA components had been discarded. By means of this filtering, the effective dimensionality of the image space had been reduced to $1,256$. Note that the patch size remains unaltered. 
Additionally, for the case with a Gamma scaling variable, the distribution of pixel intensities was rescaled (dividing by an $\alpha$ constant) such that its mean $\pm$ 3 standard deviations fit the range $[0,1]$, to have the same range for all domains. This was important for the post-training evaluations. Finally, the mean pixel intensity of each patch was subtracted before the image enters the encoder network. 

\subsubsection{NIST datasets}
We employed three popular NIST datasets: MNIST\cite{mnist}, ChestMNIST\cite{chestmnist} and FashionMNIST\cite{fashionmnist}, which contain standardized grayscale images for digits, chest radiographs and fashion items, respectively. Each original $28\times28$ pixelsize image was resized to $32\times32$ due to the specific used architecture (\Cref{subsec:methods-details}). The pixel intensities for these canonical datasets are in the range $[0,1]$.

\subsubsection{Contrast augmented NIST datasets}
Augmented versions of classic MNIST, ChestMNIST and FashionMNIST datasets were synthetically generated in order to have a rich variety in image contrast. We refer to these datasets as cMNIST, cChestMNIST and cFashionMNIST, respectively. Each image in the original dataset was first subtracted by $0.5$, so the pixel intensities move to the range $[-0.5,0.5]$. Then multiplied by a constant $c$ sampled from a Log-normal distribution, and then intensity was corrupted with a white noise vector $\boldsymbol{\eta}$, sampled from a Gaussian distribution of null mean and $\sigma_{obs}$ standard deviation

\begin{equation}
 \mathbf{x}^{aug} = c\left ( \mathbf{x}^{ori}-\frac{1}{2}  \right ) + \boldsymbol{\eta}.
\label{eqn:augmentation}
\end{equation}

\noindent
This process was repeated such that cMNIST, cChestMNIST and cFashionMNIST datasets resulting in a 10-fold increase in the size of the dataset. $\sigma_{obs}$ values for each dataset are detailed in Table S2.

\subsubsection{Shuffled pixels datasets}
For testing purposes in the out of distribution experiments, all testing subsets (Van Hateren, MNIST, ChestMNIST, cMNIST, cChestMNIST, cFashionMNIST) were used to generate synthetically pixel shuffled datasets. Given a specific testset of size $(N_{img},N_{pix})$, the pixel-shuffled dataset is generated by performing random permutation of pixels such that pixels will get permuted only with pixels that are in the same location. By doing this, the pixel-shuffled dataset would have the same likelihood of observation, yet the information present in the shuffled image will differ from the original datasets, constituting a useful baseline for out-of-distribution detection.

\subsubsection{Higher-resolution datasets} 
To evaluate uncertainty representations in more complex settings we resorted to two further datasets. Firstly,  Imagenet\cite{deng2009imagenet, russakovsky2015imagenet}, a widely used benchmark for image classification, from which we selected all three categories of human images from the standard validation set: `Ball Player', `Groom' and `Scuba Diver', as well as three object categories: `Desk', `Wall Clock' and `Bookcase'.  Secondly, we employed the NIH-ChestXray14 dataset\cite{chestmnist}. This is a large-scale public dataset of frontal chest radiographs released by the U.S. National Institutes of Health. It is composed of 112,120 images from 30,805 unique patients. The official dataset was split into 70\% to train, 10\% for validation and 20\% to test. Images were resized from $1024\times1024$ into $224\times224$, which is standard processing in medical image analysis. Same as NIST datasets, the pixel intensities for these images were set to the range $[0,1]$.

\subsection{Implementation and training details}\label{subsec:methods-details}

All results presented in the main text correspond to models that were trained in Python with Pytorch 2.x. All models were trained through Pytorch's Adam optimizer, with learning rate (lr) and weight decay (wd) as detailed in Supplementary Table S1 and Table S2, and all other Adam parameters set to Pytorch's default.  We applied early stopping so that the parameters were finally set at the epoch with best performance on the validation images. 

Following\cite{csikor2022topdown}, beta-annealing was considered to train models on natural image patches: we set $\beta_1 = 0.01$ for the first 100 epochs, linearly increasing to  the final $\beta_1 = 1$ over the next 100 epochs. We observed that in the case of the EA-VAE beta annealing had little effect in the training outcome. The final value $\beta_1 = 1$ was determined exploratorily such that the decoder's reconstructions did not present visibly strong deviations from the original input. Results are robust with respect to changes in these hyperparameters, with no abrupt qualitative changes.

\begin{figure*}[h!]
\centering
\includegraphics[width=0.8\linewidth]{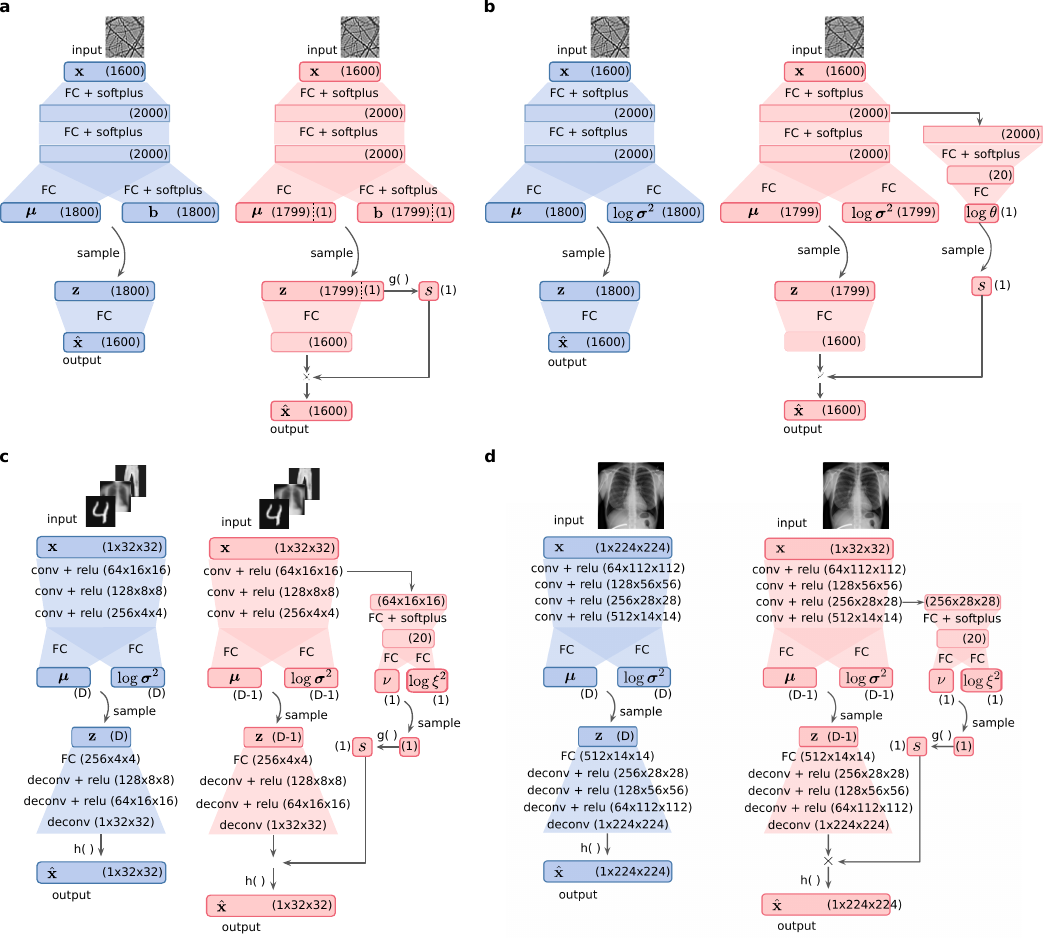}
\caption{\textbf{a}, Trained VAE (left) and EA-VAE (right) models for inference on natural images. The encoder's input of size $M=1600$ is followed by two fully connected hidden layers with softplus activation function. Two separate fully connected layers of $D=1800$ units encode the mean and variance of $q_\phi(\mathbf{z}|\mathbf{x})$. For the VAE, one fully connected layer decodes the latent sample into a reconstructed patch. For the EA-VAE model, the described VAE was modified such that the last component of the sample of the posterior distribution passes through function g() defining the effective $s$  sample from $q_\psi(s|\mathbf{x})$. The posterior sampled $s$ value is globally multiplied with the decoder's output, resulting in the final reconstructed patch. 
\textbf{b}, Trained VAE (left) and EA-VAE (right) models for inference on natural images when scaling variable assumed from a Gamma distribution. In this case, the scaling variable is inferred from a separate encoder. $k=2$ was chosen as the smallest integer number that qualitatively achieved a similar distribution to the dataset's $\sigma_{pix}$ distribution. Computationally, the choice of a fixed integer $k$, simplifies the sampling procedure, since one can always sample from the same standard Gamma function and then reparameterize by the scale parameter. This enables simple gradient computations across the sampling process.
\textbf{c}, Trained VAE (left) and EA-VAE (right) models for inference on MNIST classic and contrast-augmented domains (digits, chest x-rays and fashion). For the VAE, the encoder's input of size 32x32 is followed by three convolutional hidden layers with relu activation function. Two separate fully connected layers of $D$ units encode the mean and variance of $q_\phi(\mathbf{z}|\mathbf{x})$. A fully connected and 3 deconvolutional layers decode the latent sample. Output from last layer passes trough h() function giving place to the reconstructed image. For the EA-VAE, the described VAE model in was modified such that separate fully connected layers encode mean and variance of a Normal distribution. Its sample passes through function g() defining the effective $s$  sample from $q_\chi(s|\mathbf{x})$. The posterior sampled $s$ value is globally multiplied with the decoder's output before going through function h(), resulting in the final reconstructed image. \textbf{d}, Trained VAE (left) and EA-VAE (right) models for inference on NIH-ChestXray14 dataset. For the VAE, the encoder's input of size 224x224 is followed by four convolutional hidden layers with relu activation function. Two separate fully connected layers of $D$ units encode the mean and variance of $q_\phi(\mathbf{z}|\mathbf{x})$. A fully connected and 4 deconvolutional layers decode the latent sample. Output from last layer passes trough h() equal to sigmoid() function giving place to the reconstructed image. For the EA-VAE, the described VAE model in was modified such that separate fully connected layers encode mean and variance of a Normal distribution. Its sample passes through function g() equal to softplus() defining the effective $s$  sample from $q_\chi(s|\mathbf{x})$. The posterior sampled $s$ value is globally multiplied with the decoder's output before going through function h(), resulting in the final reconstructed image.}
\label{fig:archs}
\end{figure*}

To rule out the possibility of our findings being contingent on specific architectural choices, we explored three variants of architecture and scaling variable family for the EA-VAE, describing below the central model presented in the main text in which the scaling variable is modeled as a softplus activated Laplace distribution (See control variants details in Supplementary \Cref{subsec:implementation}). The model implements the scaling variable $s$ with minimal update to the standard VAE architecture. Here $s$ is coming from the same family of probability distributions of the regular VAE latents, $\mathbf{z}$ but transformed by a monotonous, invertible function to the positive half-space, by use of a Softplus function. The homogeneous formulation of different latent variables motivates a joint encoder for $\mathbf{z}$ and $s$, yielding a joint variational posterior $q_{\phi,\chi}(\mathbf{z},s|\mathbf{x})$. We refer to this version of the model as the \emph{homogeneous} version (see \Cref{fig:archs}a for detailed architecture).
To avoid collapse in the initial phase, a stronger regularization was applied on the scaling variable in these models for the course of the first 100 epochs ($\beta_2 = 10$), which linearly decreased to $\beta_2 = 1$ during the next 100 epochs.

To train the VAE and EA-VAE on the NIST and NIH-ChestXray14 datasets, we also incorporated convolutional layers to parametrize the encoder and decoder components, since they are commonly employed in these domains (see \Cref{fig:archs}c and \Cref{fig:archs}d for detailed architectures). In this case we used latent dimension $D$ much lower to input size, as a classic dimensionality reduction model (See Supplementary Table S2 for complete hyperparameter values).

For the contrast augmented cMNIST, cChestMNIST and cFashionMNIST models, the architectures were the same as NIST datasets, except for the last layer which lacks the sigmoid activation, corresponding to mean square reconstruction error (see \Cref{fig:archs}c for detailed architecture). $\beta_1$ and $\beta_2$ were fixed by \Cref{eqn:beta} given the observation noise in the augmentation process (See Supplementary Table S2).

All models were trained either on a GPU NVIDIA TITAN V with a 64GB RAM memory (PC: 1) or on a NVIDIA GeForce RTX 2080 Ti with a 64GB RAM memory (PC: 2). As a reference, the time taken to achieve each respective best epoch on PC 1 in this setting was of $62.2$ hours for the VAE vs $67$ hours for the EA-VAE trained on Van Hateren natural image patches. That is, the improvement in uncertainty representation in the EA-VAE comes at an expense of roughly an additional 10\% computational time. In the case of the models trained on MNIST and ChestMNIST employing convolutional layers, all models took around one hour to complete the 500 epochs.

\subsection{Post-training evaluation of VAEs and EA-VAEs}

As a general form to evaluate the models' representation of uncertainty, we computed neurons responses in the latent space for different observations. For all experiments involving uncertainty in the latent space, we define the uncertainty of the network for a given input $\mathbf{x}$ as the standard deviations of the posterior $q_\phi(\mathbf{z}|\mathbf{x})$ averaged across latent dimensions,
\begin{equation}
    u(\mathbf{x}) = \text{noise std.}=\underset{j}{\textrm{mean}} \,\, \sigma_j(\mathbf{x}).
    \label{eq:unc}
\end{equation}
We call this as well the \textit{posterior width}.

When evaluating models for inference on natural images, latent units were discerned as informative or non-informative according to their contribution to the dataset reconstruction. All experiments were computed using only the informative latent dimensions (experimental details in Supplementary \Cref{subsec:evaluation-nat}).

To assess uncertainty representations in classic computer vision domains, models were evaluated in a series of experiments that include out-of-distribution detection, image morphing and image corruption (experimental details in Supplementary \Cref{subsec:evaluation-nist}).

\section{Results}

\subsection{Inference on natural images}

To obtain a first intuition for the role of the explaining-away variable $s$, we analyzed natural image patches, where contrast is inherently present in the data, and therefore modulates uncertainty of inferences about image content (\Cref{fig:intro}f, top row and \Cref{subsec:datasets}). First, we trained both a standard VAE and an EA-VAE under identical conditions: employing the same family of distributions, data and hyper-parameters. The only difference is that in the EA-VAE we select one of the latent variables to play the role of a scaling variable, multiplicatively affecting the output. The total number of latent variables is hence the same and the model complexity does not increase in EA-VAE. 

\subsubsection{Latent representations} 
\begin{figure*}
\centering
\includegraphics[width=0.9\linewidth]{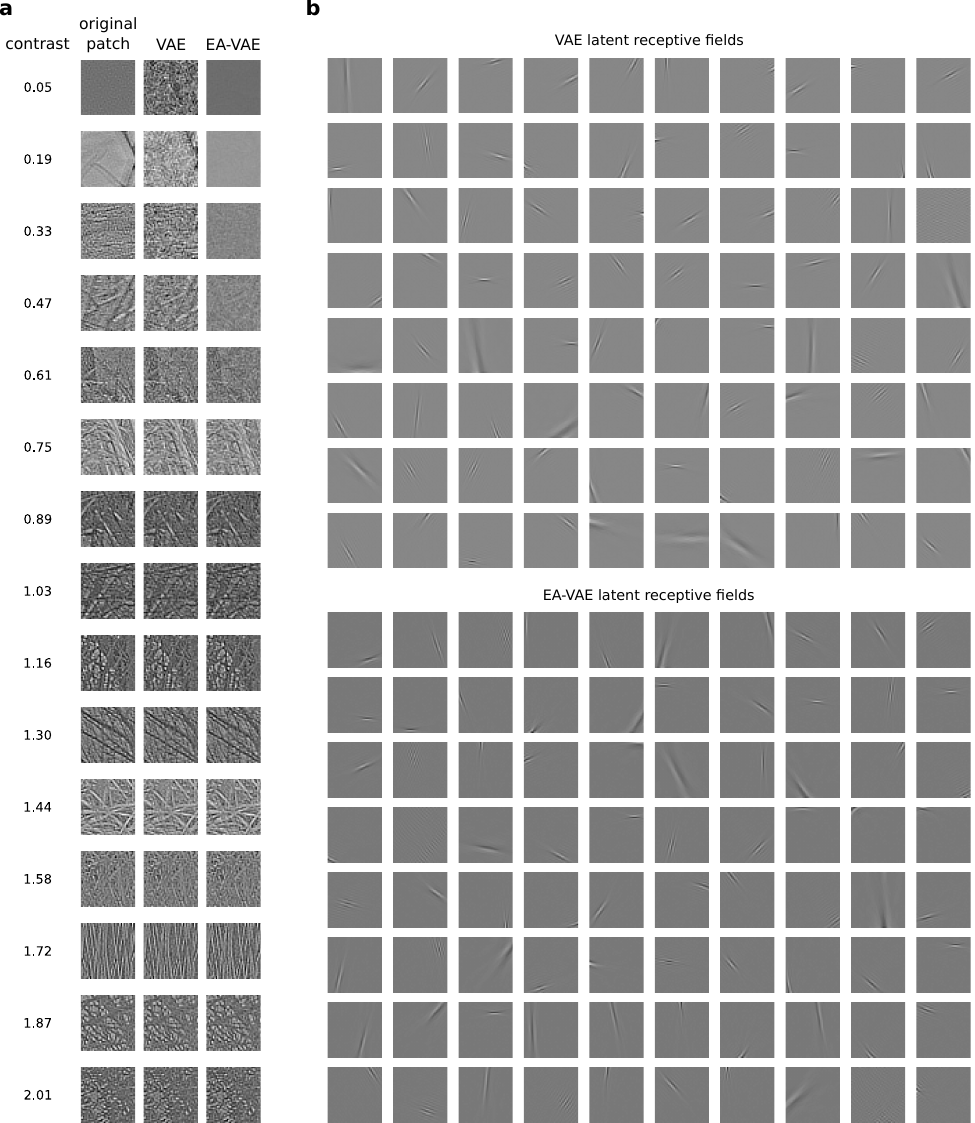}
\caption{\textbf{a}, Example test image patches of increasing contrast levels, and their respective reconstruction through the Standard VAE and the SoftLaplace-Laplace EA-VAE. \textbf{b}, Extended examples of \emph{informative} latent receptive filters in the standard VAE and the SoftLaplace-Laplace EA-VAE.}
\label{SIfig:recs_fields}
\hrulefill
\end{figure*}

We find that both the VAE and the EA-VAE can be successfully trained on this task in terms of their capacity to reconstruct natural images (\Cref{fig:intro}f and \Cref{SIfig:recs_fields}a). The models also exhibited qualitatively similar latent representations which were consistent with sparse-coding of natural images, with a fixed set of latents showing localized and oriented filters, and a smaller fraction showing unstructured filter properties (\Cref{fig:intro}g and \Cref{SIfig:recs_fields}b).

\subsubsection{Latent uncertainty} 
Characterization of latents through filter properties informs us how the \emph{mean} of the posterior is modulated by image properties. In order to characterize the relationship between how stimulus content and contrast selectively contribute to latent activations, we characterize the posteriors through the signal mean, signal variance, and noise (or posterior) variance using natural image patches that display changes in their effective contrast level. These quantities are estimated in all cases as averages over the $\mathbf{z}$ latents and over images binned according to true contrast (Supplementary \Cref{subsec:uncertainty_latent}). Both VAEs and EA-VAEs showed increased signal mean and signal variance with higher contrast, as posteriors moved away from the prior and became more distinct (\Cref{fig:intro}, cf. c and d, top and middle panels). However, in standard VAEs, posterior uncertainty (noise variance) does not decrease with increasing contrast (\Cref{fig:intro}c, bottom). Particularly, as contrast tends to zero, and local orientations become progressively harder to distinguish, posterior uncertainty fails to converge to the prior uncertainty. 
In contrast, EA-VAEs display a decreasing uncertainty for higher contrasts. Consistent with the assumptions of the generative model, as contrast decreases below the assumed pixel noise level, the reported noise variance in the EA-VAE saturates at the level of prior uncertainty (\Cref{fig:intro}d, bottom panel). This observation is also consistent with the fact that the latent $s$ is inferred as close to zero below this level (\Cref{fig:intro}h). Moreover, as expected given its multiplicative role, $s$ grows monotonically with image contrast in EA-VAEs trained on natural images (\Cref{fig:intro}h). These results suggest that the explaining away variable $s$ could be playing the role of a signal-to-noise ratio estimate, with low values corresponding to an interpretation  that the image is mostly noise, and consequently the posteriors for other latents relax towards the prior (see Supplementary \Cref{subsec:role_s}). We note that the results obtained for this homogeneous variant of the EA-VAE were qualitatively analogous to the ones obtained for the other EA-VAE variations: with the scaling variable parametrized by lognormal or gamma distributions (\Cref{SIfig:natural}).

We note that the reconstruction term of the ELBO in \Cref{eqn:VAEcost} is calculated using a single sample from the variational posterior. One may suspect this could lead to underestimation of posterior variance, hence contributing to failed inference in VAEs. We therefore investigated whether uncertainty estimates could be corrected by a VAE variant in which multiple samples are taken to evaluate the ELBO. For this, we implemented the Importance Weighted Variational Autoencoder (IWAE) \cite{burda2015importance}. In particular, we wanted to assess whether IWAEs could improve uncertainty representations in VAEs without the need to modify the generative model. We trained the IWAE on the same dataset and conditions and found no evidence of multi-sample ELBO estimates correcting the contrast-dependence of the variational posterior (\Cref{fig:iwae}).

\subsection{Improved inference of EA-VAE}

Having established a first intuition for the role of an explaining away variable in VAEs, we next explored whether the improved uncertainty representation in EA-VAEs may extend to other classical domains of computer vision. 
For this, we used standard datasets of computer vision: the MNIST, ChestMNIST, and FashionMNIST datasets, which we adapt for our investigation. To this end, we elevate these data sets with contrast-modulation (\Cref{subsec:datasets}, referred to as cMNIST, cChestMNIST and cFashionMNIST, respectively).

\begin{figure*}[t!]
\centering
\includegraphics[width=1.\linewidth]{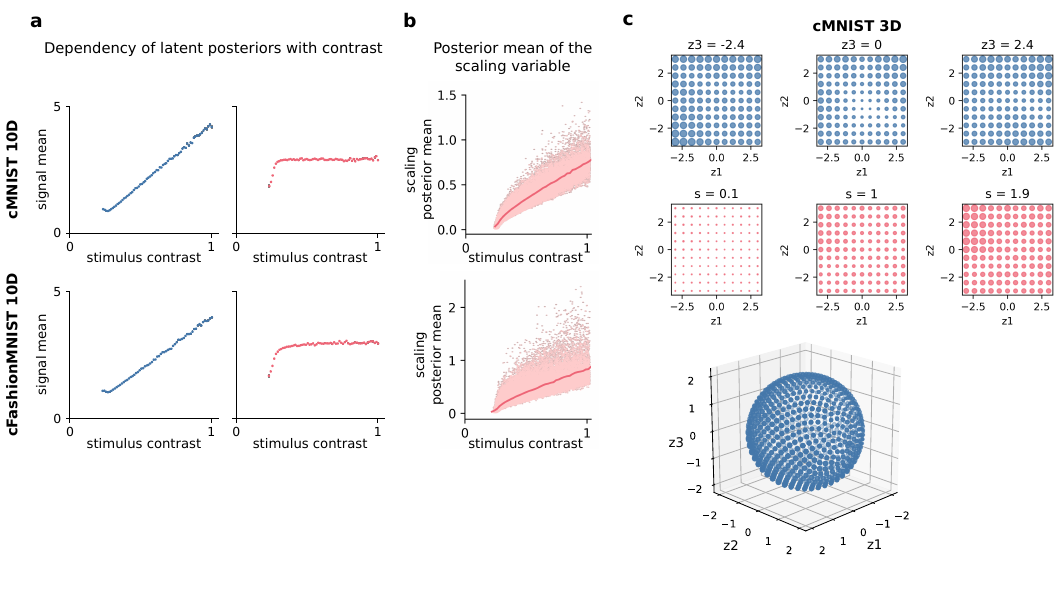}
\caption{\textbf{Characterization of learned representations for different training data sets.} Beyond natural image patches, contrast-augmented versions of the MNIST and Fashion MNIST databases are shown. \textbf{a}, Latent posterior signal mean as a function of the measured contrast of a large set of test images for both the VAE and the EA-VAE models. \textbf{b}, Inferred posterior mean of the scaling variable for individual images (\emph{dots}) in the EA-VAE model, as a function of the measured contrast of these images. Average mean shown as solid line. \textbf{c}, Direct comparison of the learned representation of a cMNIST-trained standard VAE (\emph{blue}) and an EA-VAE (\emph{red}) both featuring a three-dimensional latent space. \emph{Top panels}: Cross-sections of the latent space (see labels above panels). \emph{Dots} show the contrast of images generated from individual states of the latent space, with dot size proportional to the contrast of the generated image. Note that cross sections for the EA-VAE correspond to different levels of the scaling variable. \emph{Bottom panel}: Contrast of images generated from the three-dimensional latent space of the standard VAE model at a fixed distance from the origin. The fixed distance was 2SD of the prior distribution.}
\label{fig:rep_learning}
\hrulefill
\end{figure*}

We trained VAEs and EA-VAEs on cMNIST and cFashionMNIST datasets. In both domains we found that the signal mean grows unbounded with contrast in VAEs, while it quickly saturates for EA-VAEs (\Cref{fig:rep_learning}a). Conversely, the explaining away variable was strongly modulated by contrast in the EA-VAE for both of these extended datasets (\Cref{fig:rep_learning}b). This suggests that $s$ contributes to the disentanglement of stimulus content from contrast in EA-VAE.

Next, we sought to understand how EA-VAE's inductive bias shapes its learned representation. For this, we trained three-dimensional latent space versions of VAE and EA-VAE models, whose latent space can be directly visualized. Images generated from a grid in the latent space of the VAE show strong modulations of image contrast in all latents (\Cref{fig:rep_learning}c, top), with equal-contrast images residing approximately on a sphere (\Cref{fig:rep_learning}c, bottom). In contrast, images generated from the regular latent variables of the EA-VAE ($z_1$ and $z_2$) showed homogeneous contrast at any given value of the explaining away latent, $s$ (\Cref{fig:rep_learning}c, middle). In summary, the explaining away variable contributes to the disentanglement of contrast from other latent factors in EA-VAEs, which does not naturally occur in standard VAEs (though see \cite{higgins2017beta}).

\subsection{Out-of-distribution inferences with EA-VAE}\label{subsec:main_ood}
\begin{figure*}[t!]
\centering
\includegraphics[width=0.6\linewidth]{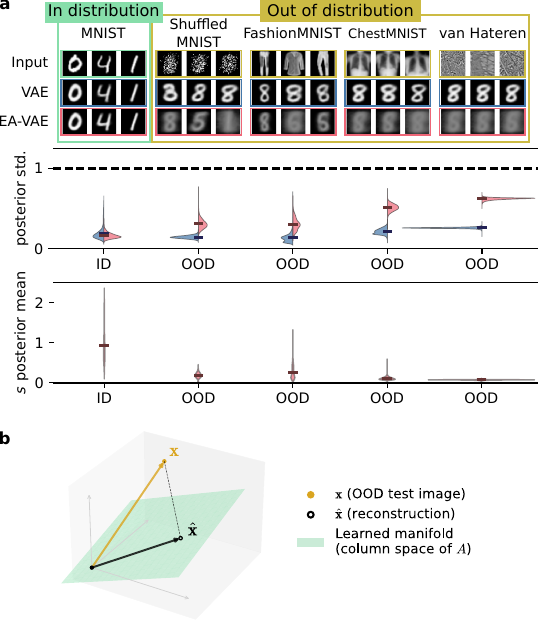}
\caption{ \textbf{Characterization of posterior width for out-of-distribution experiments in models trained on MNIST.} \textbf{a}, Posterior width and the magnitude of the scaling variable for in-distribution (\emph{ID}) and out-of-distribution (\emph{OOD}) examples both for standard VAE (\emph{blue}) and EA-VAE (\emph{red}). \emph{Top}: Examples of images from the same domain (\emph{light green frames}) and different domains (\emph{mustard frames}) as inputs, and reconstructions by the standard VAE (\emph{second row}) and EA-VAE (\emph{third row}) respectively. \emph{Bottom}: uncertainty quantified by the posterior width and mean of scaling variable posterior for individual within-distribution or out-of-distribution images. As a reference, a dashed black line indicates the prior uncertainty. \textbf{b}, Learned linear manifold spanned by in-distribution training data (green plane) and the projection $\hat{\mathbf{x}}$ of an out-of-distribution test data point $\mathbf{x}$ to the learned manifold.}
\label{fig:results-ood}
\hrulefill
\end{figure*}

After establishing that the explaining-away variable implements a direct inductive bias for learning about contrast, we next explored the contribution of this inductive bias to other computer vision tasks. In particular, VAEs were shown to struggle with detecting out-of-distribution (OOD) data, namely data points that come from a distribution whose statistics do not match the data distribution the VAE was trained on \cite{nalisnick2018deep}. Indeed, testing a VAE trained on the standard MNIST dataset shows a striking pattern: reconstructed images of a standard VAE when presenting shuffled pixels, or images coming from other datasets are surprisingly similar to in-distribution reconstructions (\Cref{fig:results-ood}a). Accordingly, posterior variances show no effect of increased uncertainty (\Cref{fig:results-ood}a). Equivalent results were found when training on the other NIST datasets (see Supplementary \Cref{SM_subsubsec:ood} and \Cref{SIfig:ood}).

\begin{figure*}
\centering
\includegraphics[width=0.45\linewidth]{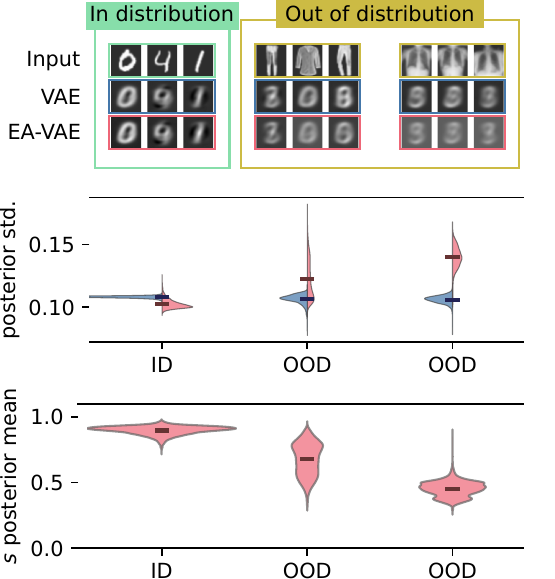}
\caption{\textbf{Out of distribution experiment when training a Standard VAE and EA-VAE on MNIST with a linear generative model.} Posterior width and the magnitude of the scaling variable for in-distribution (\emph{ID}) and out-of-distribution (\emph{OOD}) examples both for standard VAE (\emph{blue}) and EA-VAE (\emph{red}). \emph{Top}: Examples of images from the same domain (\emph{light green frames}) and different domains (\emph{mustard frames}) as inputs, and reconstructions by the standard VAE (\emph{second row}) and EA-VAE (\emph{third row}) respectively. \emph{Bottom}: uncertainty quantified by the posterior width and mean of scaling variable posterior for individual within-distribution or out-of-distribution images. Architecture of \Cref{fig:archs}c was used, without the relu activations in the decoder. Hyperparameters in Table \Cref{tab:hyperparams-mnist}.}
\label{SIfig:ood-linear}
\hrulefill
\end{figure*}

To better understand how the introduction of an explaining-away variable might address this issue, we sought to find an analytical treatment of OOD inferences in EA-VAEs. As the scaling variable $s$ is directly related to posterior uncertainty (Supplementary \Cref{subsec:role_s}), we focus on the behavior of this variable. Assuming a linear generative model where $\mathbf{A}$ maps latents ($\mathbf{z}$) to observed variables $\mathbf{x}$, it can be shown (Supplementary \Cref{subsec:ood_theory}) that the explaining-away variable in an EA-VAE can be expressed as a function of the stimulus contrast $\left\|\mathbf{x} \right\|^2$:

\begin{equation}
    s^2\big(D+\log s\big) \propto
    \left\|\mathbf{x} \right\|^2 \cos(\mathbf{x},\hat{\mathbf{x}})^2,
    \label{eq:ood_anal}
\end{equation}  

where the proportionality constant is determined by $\mathbf{A}$, $D$ is the dimensionality of the latent space, and the reconstruction $\hat{\mathbf{x}}$ corresponds to the projection of $\mathbf{x}$ to the learned linear manifold spanned by in-distribution data (\Cref{fig:results-ood}b). Importantly, the second factor on the right hand side of Eq.~\ref{eq:ood_anal} is the cosine similarity between the input data points and the corresponding reconstructed data points. This provides an intuitive insight: when presenting an image that is an outlier to the learned manifold spanned by in-distribution data, the cosine similarity will be low and hence so will be $s$. In turn, lower $s$ values encourage posteriors to collapse to the prior for those inputs, increasing posterior variances (Supplementary \Cref{subsec:role_s}). In summary, the EA-VAE is expected to display higher variance for OOD images. An EA-VAE that features a linear decoder confirms these analytical results (\Cref{SIfig:ood-linear}), and the general setting that features a nonlinear decoder shows that, in contrast with the standard VAE, the uncertainty of inferences in the EA-VAE is systematically modulated by mismatched training and test statistics (\Cref{fig:results-ood}a).

\begin{figure*}[t!]
\centering
\includegraphics[width=0.5\linewidth]{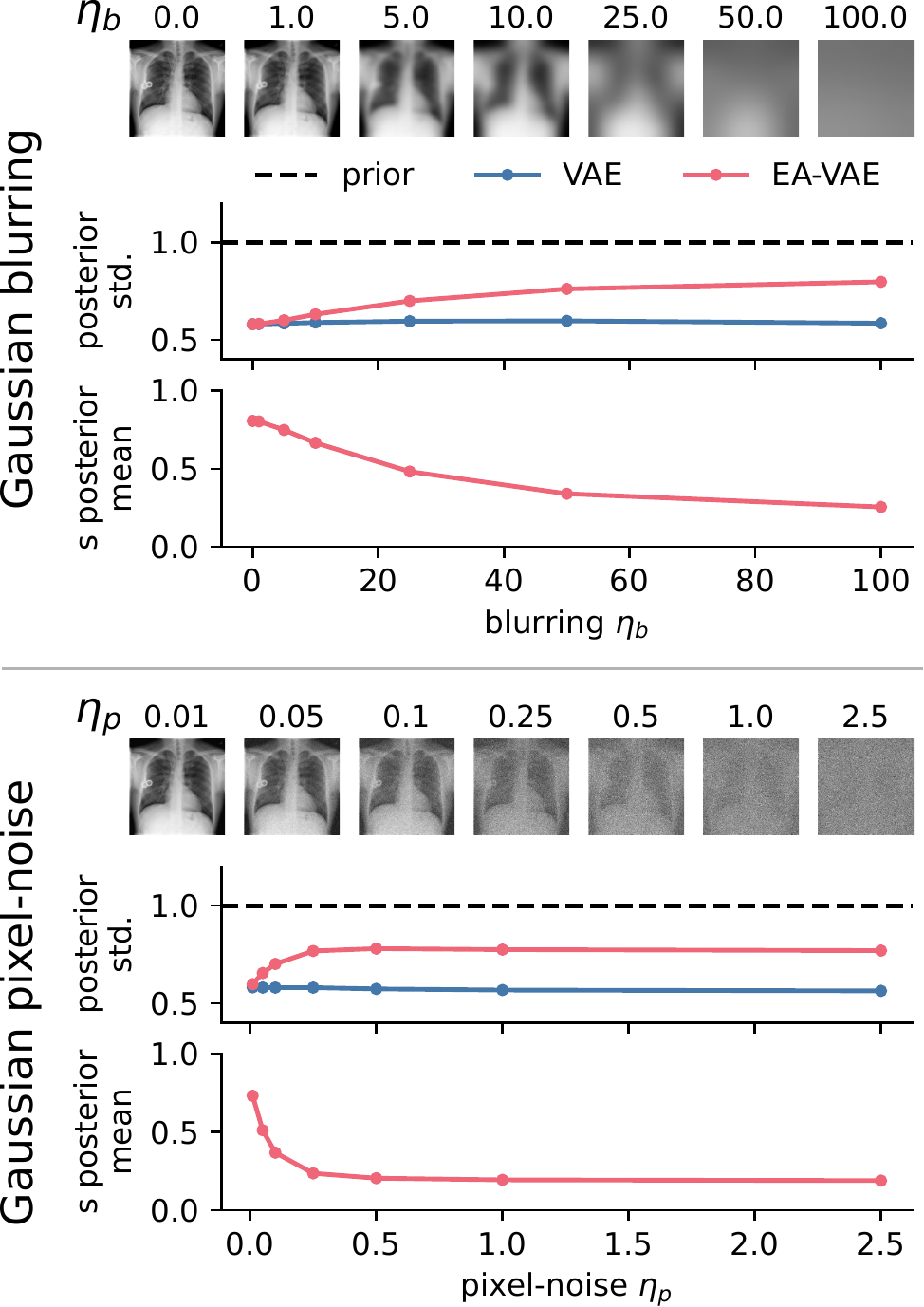}
\caption{\textbf{Uncertainty representations in high-dimensional medical images.} Posterior width and magnitude of the scaling variable of standard VAE and EA-VAE models trained on NIH-ChestXray14, for images increasingly corrupted either by Gaussian blurring (top) or additive pixel-noise (bottom).}
\label{fig:results-corruption}
\hrulefill
\end{figure*}

\begin{figure*}[t!]
\centering
\includegraphics[width=0.45\linewidth]{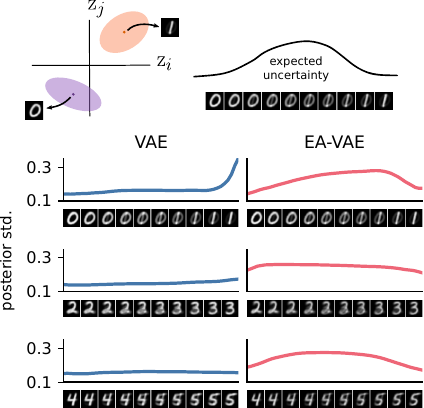}
\caption{ \textbf{Characterization of posterior width for morphing experiment that affect inference uncertainty for models trained on MNIST.} \emph{Top left}: Average posterior for images with the same label ('0' in \emph{purple} and '1' in \emph{orange}). \emph{Top right}: Gradual morphing between  representative image samples from digit '0' and digit '1'. Illustration of the expected changes of the posterior width is shown above. \emph{Bottom}: Posterior widths for the morphed digits with standard VAE (\emph{left}) and EA-VAE (\emph{right}).}
\label{fig:results-morphing}
\hrulefill
\end{figure*}

Inspired by this insight, and encouraged by the results obtained on simple MNIST datasets, we tested whether the ability of EA-VAEs to detect OOD examples could extend to a more complex setting such as distinguishing human versus non-human images from Imagenet \cite{deng2009imagenet, russakovsky2015imagenet} (Supplementary \Cref{subsec:inception}). Concretely, we trained VAEs and EA-VAEs on images from one human category and evaluated them on images either from another human category or on images from an object category. Interestingly EA-VAEs showed more dissimilar distributions of uncertainty between human and object images than VAEs, hinting at a more structured semantic representation of uncertainty (\Cref{fig:inception}). These results hint at a promising avenue of research into OOD detection with EA-VAEs in broader more complex settings.

\subsection{Image corruption and uncertainty}\label{subsec:maincorruption}

\begin{figure*}
\centering
\includegraphics[width=0.5\linewidth]{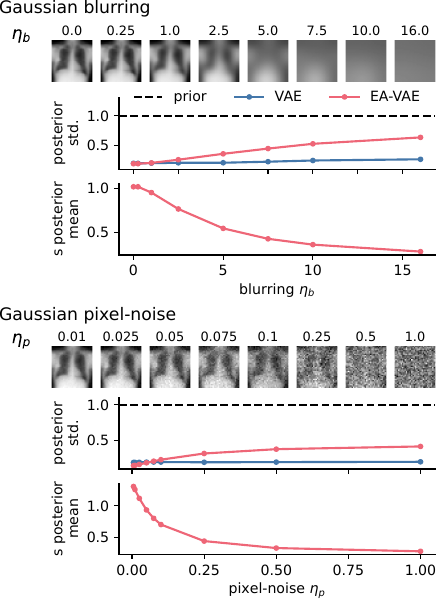}
\caption{Posterior width and the magnitude of the scaling variable of standard VAE and EA-VAE models trained on ChestMNIST, for images increasingly corrupted either by Gaussian blurring (top) or additive pixel noise (bottom).}
\label{SIfig:corrupt-chest}
\hrulefill
\end{figure*}

OOD experiments confirmed that the scaling variable introduced in EA-VAE enables it to reflect the mismatch between the training data and the input through posterior uncertainty. Another form of relevant mismatch to be flagged by models concerns images of lower quality or subjected to corruption that may impede correct inferences. To assess this, we analyzed the representation of uncertainty in VAEs and EA-VAEs trained on Chest x-ray images. In order to showcase the scalability of EA-VAEs to larger scale problems we present here the results corresponding to higher resolution images from the NIH-ChestXray14 (\Cref{subsec:datasets}), though equivalent results were found for the lower definition images from ChestMNIST employed in the OOD analysis (\Cref{SIfig:corrupt-chest}).

We progressively corrupted images, rendering them increasingly uninformative, in two ways: by applying a Gaussian blur (\Cref{fig:results-corruption} top) and by adding Gaussian pixel noise (\Cref{fig:results-corruption} bottom) (Supplementary \Cref{subsec:image_corruption}). In both cases the EA-VAE displays increasing uncertainty as the image quality deteriorates. This starkly contrasts with the behavior of standard VAEs, where latent uncertainty remains almost constant at low values even as the original image is no longer recognizable from the corrupted one (cf. blue and salmon lines, \Cref{fig:results-corruption}). This behavior was consistent with the observation that, when faced with an intentionally uninformative image obtained by averaging images over the entire dataset, EA-VAEs report high uncertainty, while standard VAEs do not (\Cref{SIfig:uninf-mnist}). 

\subsection{Morphing and uncertainty}\label{subsec:main_morphing}

\begin{figure*}[t!]
\centering
\includegraphics[width=1.0\linewidth]{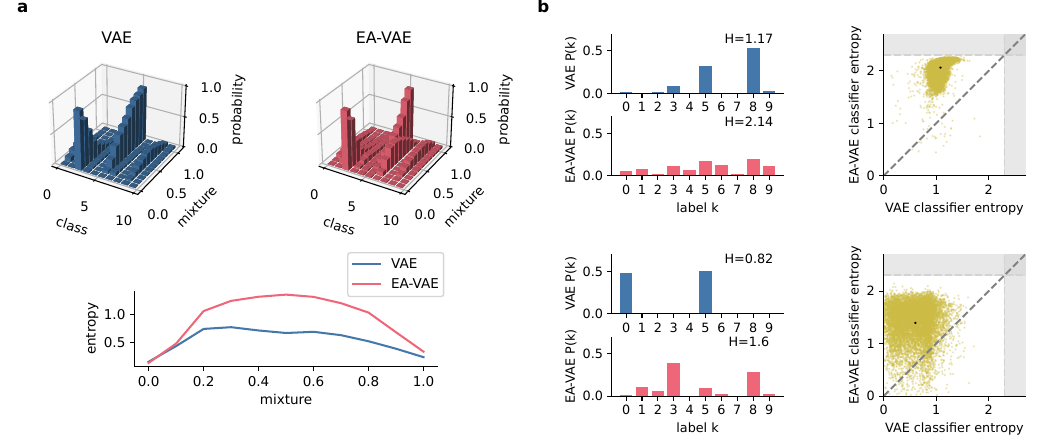}
\caption{\textbf{Predictive uncertainty in a downstream classification task in VAEs and EA-VAEs.} Characterization of an MLP trained for classification task of MNIST handwritten digits through latent posterior representations of a VAE and an EA-VAE. \textbf{a}, Behavior of the trained MLP when gradually morphing between image samples from digit '2' to digit '5'. \emph{Top panels}: Histograms of output predicted probabilities averaged for different samples of digits '2' and '5' when trained with standard VAE (left) and EA-VAE (right) latent representations. \emph{Bottom panel}: Entropy of predicted probability distribution as a function of combination weight of labels. \textbf{b}, Behavior of trained MLP when testing with out of distribution ChestMNIST (\emph{Top}) and pixel-shuffled MNIST (\emph{Bottom}) images. \emph{Left columns}: Histograms show the distribution of predicted probabilities of labels of an example image, averaged for different samples of the posterior distribution of that image. The entropy of that averaged distribution is computed for both models. \emph{Right column}: Entropy of predicted probability distribution for an MLP trained with standard VAE representation vs EA-VAE representations. Each yellow dot represents a single x-ray image. Black dot in mean value. Identity in dark gray slashed line. Shading corresponds to the upper limit determined by the uniform distribution.}
\label{fig:class_task}
\hrulefill
\end{figure*}

When dealing with a domain containing distinct stimulus categories, the latent space is expected to adopt a non-overlapping representation in which inferences about individual images of a given category yield a low uncertainty estimate after learning. In contrast, inference is expected to reflect an increased uncertainty for images that might belong to multiple categories. Note that as VAEs are trained in an unsupervised fashion, these do not rely on category labels during training. Still, we argue that distinct categories correspond to learned subdomains of the image space and morphing systematically drives images to domains of the pixel space with more limited exposure during training. Systematic changes in uncertainty due to ambiguous image categories can be investigated in the case of MNIST-trained models by assessing posterior uncertainty for images of simple digits and for images that are produced by merging two digits from different classes. We conducted this numerical experiment for each of the 45 pairs of different digits and measured the uncertainty along different levels of interpolations between the digits (\Cref{fig:results-morphing}, and \Cref{SIfig:morphing}, for alternative pairings). We found a consistent modulation of uncertainty for the EA-VAE model, featuring a clear maximum at an intermediate interpolation level in all except one of the digit pairs (see also Supplementary \Cref{subsec:morphing}). In contrast, no consistent peak was identified for the standard VAE model: only 31\% of all cases featured a central maximum (14 out of 45 in total).

Methods such as Monte-Carlo (MC) Dropout \cite{gal2016dropout}, have often been proposed to achieve better uncertainty estimates in deep learning, by keeping drop-out layers active at inference time and incorporating the additional variance introduced by drop-out. As a control experiment, we studied whether MC Dropout would suffice to remedy uncertainty representations without affecting the generative models. We repeated both morphing and out-of-distribution experiments and found this did not correct uncertainty estimates in these cases (\Cref{fig:dropout}). Controls such as this and that of IWAEs provide further support for the idea of a fundamental mismatch in the basic assumptions VAEs make about the statistics of the data, which require addressing how the generative model itself is modeled.

\subsection{From latent to predictive uncertainty}

Next, we investigated whether the benefits from improved latent uncertainty representations in EA-VAEs also transferred to improved predictive uncertainty in a downstream classification task. Optimal information fusion relies on computing with proper uncertainty estimates, and VAEs are often employed for dimensionality reduction as a previous step to classification. Hence, when multiple sources need to be combined, as it is often the case in medical settings, improved uncertainty estimates could be beneficial.

\begin{figure*}
\centering
\includegraphics[width=0.9\linewidth]{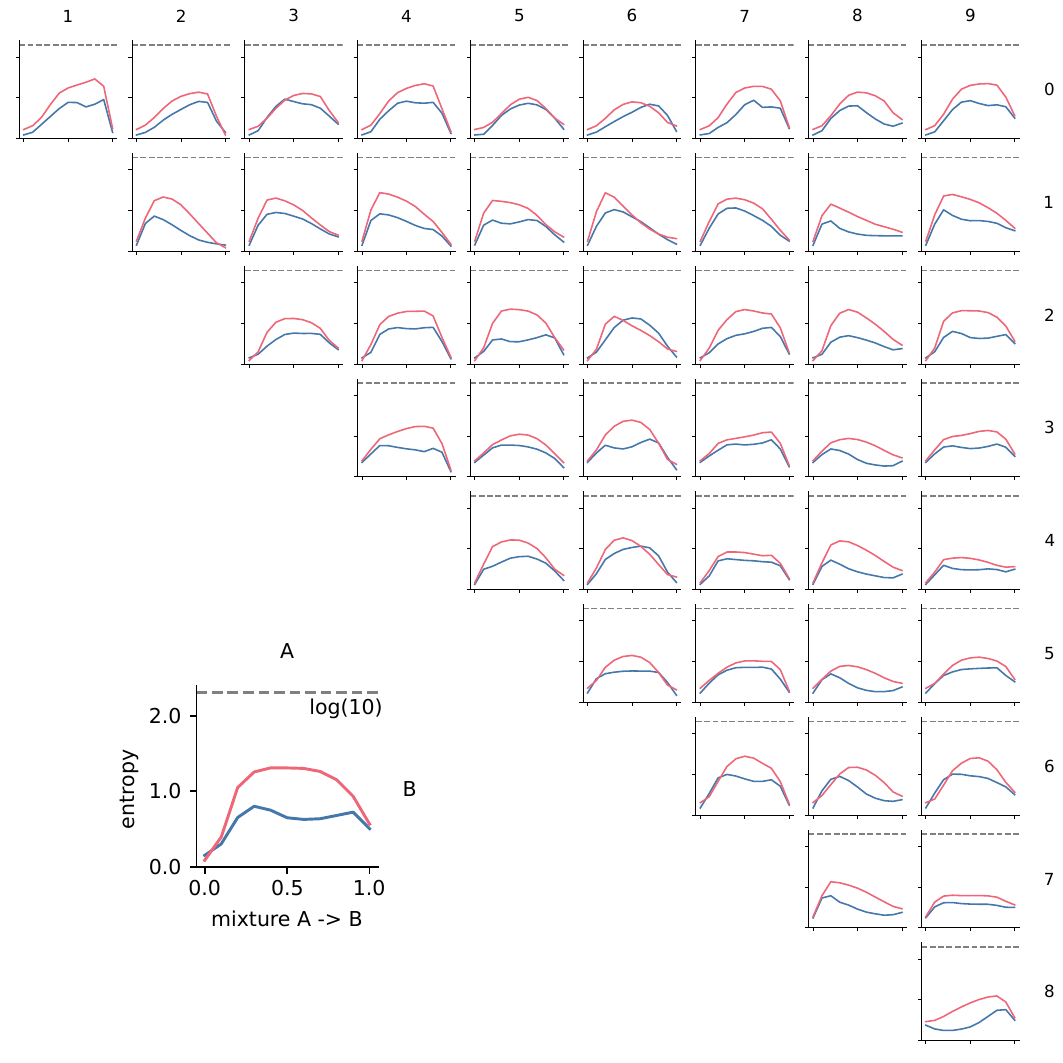}
\caption{Behaviour of trained MLP classifier when gradually morphing between image samples from digit 'A' to digit 'B' when trained with standard VAE (blue) and EA-VAE (red) latent representations. Entropy of predicted probability distribution as a function of combination weight of labels for all 45 possible combinations of labels. Dashed line corresponds to the upper limit determined by the discrete uniform distribution.}
\label{SIfig:entropy}
\hrulefill
\end{figure*}

To investigate this, multilayer perceptrons (MLPs) were trained to classify MNIST digits using samples from the VAE (or EA-VAE) posterior latent representations (Supplementary \Cref{subsec:mlp_classifier}). 
Two experiments evaluated the impact of these representations on information fusion and out-of-distribution examples. First, using a set of morphed digits, we compared the predictive uncertainty of MLPs, quantified as the entropy of the output distributions. The EA-VAE demonstrated significantly higher uncertainty (entropy) for intermediate images compared to the VAE ($p_{\text{value}}< 10^{-15}$). (\Cref{fig:class_task}a and \Cref{SIfig:entropy}) 
Second, testing on out-of-distribution inputs (ChestMNIST and pixel-shuffled MNIST), the EA-VAE again yielded higher entropy predictions than the VAE (\Cref{fig:class_task}b).

\begin{figure*}[t!]
\centering
\includegraphics[width=1.0\linewidth]{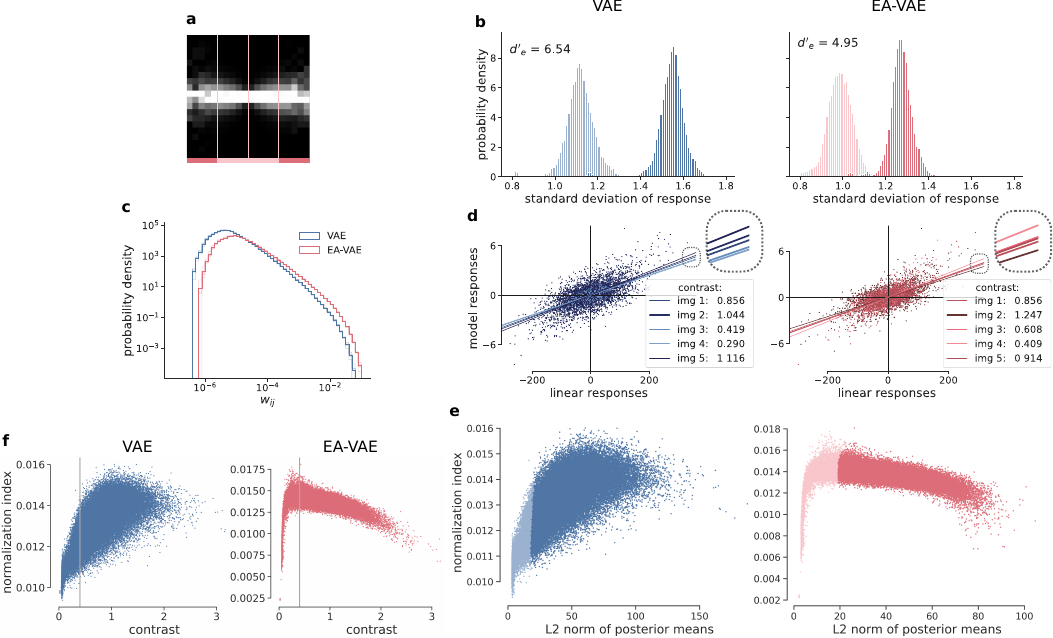}
\caption{\textbf{Divisive normalization implemented by the recognition model.} {\bf a}, Conditional distribution ($p(L_1\,|\, L_2,x)$) of linear filter responses of a pair of example latent variables over natural images. Intensity of histogram is proportional to the probability. Linear filter responses are calculated as dot product between an image and the receptive field of the neuron (Supplementary \Cref{subsec:div_norm}). \emph{Vertical} and \emph{horizontal} axes correspond to $L_1$ and $L_2$ response intensities. \emph{Light and dark colored bars} correspond to two central and two flanking quadrants of $L_2$ activation, respectively.  {\bf b}, Non-linear dependence of latent posteriors. Dependence is characterized by 100 randomly chosen pairs of latents using natural image posteriors. Standard deviation of the conditional distribution of posterior means are shown close to zero activation of the conditioned latent variable (two central quadrants, marked with matching color as panel a) and at high-intensity activation (two flanking quadrants, color as on a). {\bf c}, Distribution of learned weights of the divisive normalization model for the standard VAE (\emph{blue}) and EA-VAE (\emph{red}) models. \emph{Dark line} shows the average of five fits, individual fits are shown with \emph{light lines}. The two distributions differ significantly (Kolmogorov-Smirnov test, $D = 0.598, p<.001$). {\bf d}, Evaluation of divisive normalization  through contrasting linear responses with posterior means in the standard VAE (\emph{blue}) and EA-VAE (\emph{red}) models. \emph{Dots} represent responses of individual latents for a particular natural image, \emph{colors} distinguish responses to individual natural images, color lightness set according to image contrast. \emph{Lines} show linear fits to the responses. Slope of the fit is designated as the normalization index, such that a smaller index corresponds to a decreased dependence of the posterior on the linear response, i.e. stronger divisive normalization. Normalization indices, and contrasts of the five example images are shown in the \emph{legend}. {\bf e}, Normalization index in the standard VAE (\emph{blue}) and EA-VAE (\emph{red}) models as a function of the normalization factor of the divisive normalization model, i.e. the Euclidean mean posterior means. Dots correspond to individual images, light colors correspond to images with contrast levels below observation noise (Supplementary \Cref{subsec:div_norm}). {\bf f}, Normalization index in the standard VAE model (\emph{blue}) and EA-VAE model (\emph{red}) as a function of image contrast. \emph{Dots} represent the normalization index for a particular natural image. \emph{The vertical line} signals the set observation noise. 
\label{fig:dn}}
\hrulefill
\end{figure*}

These results confirm that differences in the properties of latent uncertainty representations between VAEs and EA-VAEs also translate to differences in output uncertainty, when those representations are used downstream for a supervised task.

\subsection{Signatures of divisive normalization in the recognition model}

So far, we have demonstrated that the inductive bias in the generative model contributes to more accurate inference. Next, we investigated how the  recognition model is adapted to the inductive bias in the generative model. This ultimately promises to contribute to understanding the computational implementation that enables EA-VAEs to accomplish improved inference. 

Contrast-related changes induce a dependency in the joint statistics of linear filters, referred to as bow-tie dependency: pairwise conditional distributions display a characteristic bow-tie shape (\Cref{fig:dn}a). As VAEs assume independence of latent variables, we draw inspiration from neuroscience, where divisive normalization was proposed to reclaim independence \cite{schwartz2001natural}. 

We trained the divisive normalization model (\Cref{eq:dn}, Supplementary \Cref{subsec:div_norm}) on both the standard VAE and the EA-VAE. As the scaling variable explains contrast-related changes, we expected that the bow-tie dependency is reduced in EA-VAE. To analyze this, we computed the conditional distributions of the responses of latent pairs over natural images (\Cref{fig:dn}a) and calculated the width of the conditional distribution both at low-, and high-level activations of the conditioned latent (\Cref{fig:dn}b). High-level activations were characterized by higher variance in both the standard VAE and EA-VAE, but the difference was reduced in the EA-VAE (\Cref{fig:dn}b). We found that the reduced bow-tie dependence could be explained by stronger divisive normalization in the EA-VAE, as reflected by stronger weights learned by the divisive normalization model (\Cref{fig:dn}c).

We also assessed divisive normalization by measuring the degree of suppression of individual latents depending on the activity of the remaining latent variables. Suppression was measured through the normalization index, which is based on the relationship between the linear responses and posterior means: a small normalization index indicated stronger suppression relative to the response expected from a linear model (\Cref{fig:dn}d). We compared the response intensity of the latents with the normalization index for both the EA-VAE and standard VAE models (\Cref{fig:dn}e). While the EA-VAE displayed consistently higher normalization with more intense latent activations, the standard VAE showed an opposite tendency, indicating that divisive normalization was characteristic of EA-VAE but not the standard VAE. Note that observation noise limits the quality of inference, as images with contrast lower than the level of observation noise are considered merely as noise by the generative model. Increased contrast results in higher signal mean (\Cref{fig:intro}c-d) and therefore contrast might have a direct effect on the strength of divisive normalization. Indeed, contrast-dependence of the normalization index identified a strong effect of divisive normalization in the EA-VAE model, but not in the standard VAE (\Cref{fig:dn}f).

In summary, we identified signatures of divisive normalization in EA-VAEs, supporting the conclusion that the inductive bias shapes the learning of the recognition model of the EA-VAE.

\section{Discussion}

Deep generative models are praised for establishing a coherent framework for probabilistic machine learning \cite{ghahramani2015, Goodfellow2016}, and VAEs are particularly appealing as they provide a normative tool for learning of (and inference in) generative models of arbitrary complexity \cite{kingma2019}. At the same time, the capacity of popular generative methods, and VAEs in particular, to faithfully represent the uncertainty of inferences has been put to question \cite{nalisnick2018deep}. This makes the systematic study of uncertainty representations in VAEs key to know when not to trust them, and find a potential way forward to improve them. Indeed, our investigations consistently revealed pervasive problems in uncertainty representation of VAEs, which we argued stems from ignoring non-linear statistical dependencies characteristic of natural and artificial images. Motivated by earlier computer vision insights that addressed these non-linear dependencies through multiplicative latent variables as in the Gaussian Scale Mixture (GSM) model \cite{NIPS1999_6a5dfac4}, we propose a novel framework that combines the simplicity and elegance of the VAE formulation with the flexibility of scale mixture models. In this model family, which we call EA-VAEs, the joint activation of a large number of local variables in the VAE can be \emph{explained away} by an increase in the global multiplicative variable. This allows EA-VAEs to effectively capture non-linear dependencies in the data. Critically, we show that while the inclusion of this explaining away variable was initially motivated by the need to adequately model contrast-related changes in the uncertainty of inferences over natural images, the benefits of the framework extend to a variety of issues in a wide array of computer vision domains. This includes the highly relevant issues of uncertainty quantification when faced with out-of-distribution, or corrupted data. Interestingly, our mathematical analyses of the EA-VAE model provided valuable insights into the geometrical properties of its learned representations, how these are fostered during training, and the corresponding mechanistic underpinnings of the solution. Indeed, our analyses revealed that along with the inductive bias introduced in the generative model, the recognition model of the VAE also underwent adaptation, developing signatures of divisive normalization, a widely reported motif in biological neural networks \cite{carandini2012normalization}. These results highlight the benefit of neuroscience-inspired ideas in artificial intelligence \cite{Sinz2019}.

The proposed extension of the standard VAE model requires a small modification to the architecture, at no significant additional computational cost. We show however, that this simple intervention has a strong impact in the models' inductive biases, starkly improving on standard VAEs' capabilities. We argue that it is a principled extension because it directly addresses a common assumption in VAEs: the independence of latent variables. As argued, this assumption is easily violated. While this could in principle be addressed by learning a different representation to accommodate this dependency through strong nonlinearities in the generative model of VAEs, this does not occur in practice, where directly eliminating this source of nonlinearity appears to provide a more viable path. We have indeed investigated alternative strategies to remedy uncertainty representations that do not involve changes to the generative model. Specifically, we have explored MC dropout and IWAES, which tackle two potential sources of uncertainty underestimation stemming from epistemic uncertainty and from the use of a single sample from the variational posterior to estimate the loss in VAEs. We found no evidence that either of these techniques could restore effective inference.

Moreover, we argue that the connection between the explaining away phenomenon in probabilistic inference and divisive normalization as a computational motif should be further explored and exploited in machine learning. As shown in Ref.~\cite{echeveste2020cortical}, artificial neural systems optimized for full sampling-based Bayesian inference for models with explaining away variables result in solutions with mean responses which precisely recover divisive normalization. Our work  provides evidence for this link in a variational setting, hinting at a wider pattern. It is hence possible that, as with divisive normalization in the brain, explaining away variables in artificial neural architectures may play an important role in multiple domains. 

We have taken a thorough and layered approach to our study of EA-VAEs: building from first intuitions into the role of explaining away variables, which were then supported by analytical derivations and toy examples, escalating until reaching complex higher dimensional datasets. Throughout these cases we have systematically shown the beneficial properties of EA-VAEs. Inference models which can readily provide a meaningful representation of uncertainty are crucial in a plethora of fields, such as in the healthcare domain, where optimal information fusion relies on weighting multiple information sources by their relative uncertainties \cite{ricci2023towards}. While further work is required to assess these benefits in other non-visual modalities, we believe EA-VAEs constitute a promising model type for a wide range of scenarios. Finally, the simplicity and robustness of the proposed updated architecture promises effortless adoption of the EA-VAE architecture in practical applications.

\section{Data and code availability}
All datasets used in this paper are publicly available. Van Hateren dataset of natural image patches is provided in Ref.~\cite{csikor2022topdown}. MNIST and FashionMNIST datasets can be downloaded via Pytorch. ChestMNIST and NIH-ChestXray14 datasets can be downloaded and imported using the MEDMNIST+ official repository. Inception pre-trained network weights can be imported with Pytorch, and ImageNet database can be downloaded from the official webpage. 

Implementation code and instructions for reproducing results are freely available on GitHub, \url{https://github.com/josefina-catoni/EA-VAE-uncertainty-in-latent-representations}, \url{https://github.com/martosdomo/EA-VAE-natural-domain}.

\section{Acknowledgments}
This work was supported by Argentina’s National Scientific and Technical Research Council (CONICET) (RE), which covered the salaries of all the authors in Argentina, the European Union project RRF-2.3.1-21-2022-00004 within the framework of the Artificial Intelligence National Laboratory (GO) and the National Research, Development and Innovation Office under grant agreement ADVANCED 150361 (GO). The authors gratefully acknowledge NVIDIA
Corporation for the GPU computing, and the support of Universidad Nacional del Litoral (Grants CAID-PIC-50220140100084LI,
50620190100145LI), ANPCyT (PICT-PRH-2019-00009, 2022-03-00593).


\putbib 
\end{bibunit}
\vfill
\pagebreak

\clearpage
\begin{center}
\Huge Supplementary Material for: \\
\LARGE Remedying uncertainty representations in visual inference through Explaining-Away Variational Autoencoders \\
\large Josefina Catoni, Domonkos Martos, Ferenc Csikor, Enzo Ferrante, Diego H. Milone, Balázs Meszéna, Gergő Orbán, and Rodrigo Echeveste
\end{center}
\beginsupplement

\begin{bibunit}
\section{Supplementary Theoretical Analysis and Derivations}\label{subsec:theoretical}

\subsection{On the role of the explaining-away variable \texorpdfstring{$s$}{s}}\label{subsec:role_s}

Let us consider an EA-VAE with spatial latents $\bf{z}$ and an explaining-away latent $s$. The output, or reconstruction, of the EA-VAE can be expressed in the general form:
\begin{equation}
\hat{\mathbf{x}} = s \, \mathbf{f}(\mathbf{z}) + \mathbf{b},
\label{seqn:EAVAErec_general}
\end{equation}
where we have included an eventual constant bias term $\mathbf{b}$. The cost function of the EA-VAE, as for the VAE, can be written in terms of a reconstruction term and a regularization term which creates a pull towards the latent priors:
\begin{align}
\mathcal{L}_{EA-VAE} = \mathcal{L}_{Rec}(\mathbf{x},\hat{\mathbf{x}}) + \mathcal{L}_{Prior}(s,\mathbf{z}),
\label{seqn:EAVAEcost_general}
\end{align}
where we have made explicit that the reconstruction loss depends on the input and its reconstruction, while the regularization term is a function of the latent variables.
We are interested in understanding the role of $s$ beyond the use case of image contrast, and why one could expect $s$ to function as a general modulator of uncertainty. In particular, we have observed throughout the presented experiments that $s=0$ indicates an unknown input, accompanying maximal posterior uncertainty (and therefore a collapse towards the prior) for $\bf{z}$.

Let us note then that when $s = 0$, because of the multiplicative EA-VAE formulation, the output $\hat{\bf{x}}$ in \Cref{seqn:EAVAErec_general} becomes independent of the spatial latents $\bf{z}$. Let us recall that we have denoted in the main text as $\phi$ the parameters of the encoder for $\bf{z}$, and as $\chi$ the parameters for $s$. When $s = 0$, the gradient of the loss with respect to $\phi$ becomes:
\begin{align}
\nabla_\phi \mathcal{L}_{EA-VAE} = \nabla_\phi \mathcal{L}_{Rec}(\mathbf{x},\hat{\mathbf{x}}) + \nabla_\phi \mathcal{L}_{Prior}(s,\mathbf{z}) = \nabla_\phi \mathcal{L}_{Prior}(s,\mathbf{z}),
\label{seqn:EAVAEgradient_s_0}
\end{align}
since the first term vanishes, as the reconstruction loss will be independent of $\bf{z}$ (and hence of its associated parameters $\phi$). This means that gradient updates will modify $\phi$ in the direction:
\begin{align}
\dot{\phi} \propto -\nabla_\phi \mathcal{L}_{EA-VAE} = -\nabla_\phi \mathcal{L}_{Prior}.
\label{seqn:EAVAEstep_s_0}
\end{align}

We note that for those examples for which $s = 0$, the gradient update rule will pull the parameters $\phi$ of the network only in the direction to minimize the distance to the prior. Latent $s$ hence acts as a simple straightforward selector for the network, determining which inputs the model will try to reconstruct and for which a collapse to the prior is optimal. In this way, the EA-VAE provides a natural and simple solution to the representation of uncertainty for unknown inputs.

\subsection{On the relationship between the scale variable and out-of-distribution inputs}\label{subsec:ood_theory}

In this analytic treatment we investigate how the latent uncertainty and scaling variable in EA-VAE is affected if the model is presented with an Out-of-Distribution test data point. We are considering a simplified setting, in which the generative model component is linear. In such a case we assume $\textbf{A}$ is learned in the training phase and in-distribution images are lying on the plane defined by $\textbf{A}\cdot \mathbf{z}$. While observation noise can explain deviations from this plane, data that are farther away from the plane are less likely to be compatible with the model. For simplicity, we are performing the analysis for prior and variational posterior for $\vect{z}$ that are normal-distributed.

The scale variable $s$ is modeled with a lognormal approximate posterior,
\begin{equation}
q(s\mid \mathbf{x})=\mathrm{LogNormal}(\nu(\mathbf{x}),\xi^2(\mathbf{x})),
\qquad \log s\sim\mathcal N(\nu(\mathbf{x}),\xi^2(\mathbf{x})),
\end{equation}
and a lognormal prior with unit variance. We make the simplifying assumption that the posterior standard deviation of $\log s$ is negligible,
\begin{equation}
\xi(\mathbf{x})\approx 0,
\end{equation}
so that $s$ can be treated as deterministic and equal to its posterior mean parameter,
\begin{equation}
s=\exp(\nu(\mathbf{x})).
\end{equation}
Under this assumption, the KL divergence for $s$ contributes a quadratic penalty $\tfrac12(\log s)^2$ (up to constants).

In this case, the loss function (negative ELBO) can be written as
\begin{equation}
    \mathcal{L}(\vect{x}) =
\mathbb E_{q(\vect{z},s\mid \vect{x})}\!\left[\frac{1}{2\sigma_{obs}^2}\norm{\vect{x}-A(s\vect{z})}^2\right]
+
\mathrm{KL}\!\left(q(\vect{z}\mid \vect{x})\,\|\, p(\vect{z})\right)
+
\frac12(\log s)^2.
\end{equation}
Reparametrizing the variational posterior with its mean yields
\begin{equation}
    \vect{z}=\boldsymbol{\mu}(\vect{x})+\boldsymbol{\varepsilon},
    \qquad
    \boldsymbol{\varepsilon}\sim\mathcal N\!\big(\vect{0},\diag(\boldsymbol{\sigma}^2(\vect{x}))\big),
\end{equation}
where $\boldsymbol{\mu}(\vect{x})$ and $\boldsymbol{\sigma}(\vect{x})$ are the mean and variance of the variational posterior. For notational simplicity, we omit the dependence of these on the image. With this, the reconstruction term becomes
\begin{eqnarray}
    \mathbb E_{\varepsilon}\!\left[\frac{1}{2\sigma_{obs}^2}\norm{\vect{x}-\vect{A}(s\boldsymbol{\mu}+s\boldsymbol{\varepsilon})}^2\right] 
    &=& \frac{1}{2\sigma_{obs}^2}\norm{\vect{x}-\vect{A}(s\boldsymbol{\mu})}^2
    +
    \frac{s^2}{2\sigma_{obs}^2}\,\mathbb E_{\varepsilon}\norm{A\varepsilon}^2 \nonumber \\
    &=&
    \frac{1}{2\sigma_{obs}^2}\norm{\vect{x}-\vect{A}(s\boldsymbol{\mu})}^2
    +
    \frac{s^2}{2\sigma_{obs}^2}\,
    \mathrm{tr}\!\big(\vect{A}\,\diag(\boldsymbol{\sigma}^2)\,\vect{A}^\top\big) \nonumber \\
    &=&
    \frac{1}{2\sigma_{obs}^2}\norm{\vect{x}-\vect{A}(s\boldsymbol{\mu})}^2
    + \frac{s^2}{2\sigma_{obs}^2}\,
    \sum_{j=1}^D v_j\,\sigma_{j}^2,
\end{eqnarray}
where we introduced $v_j := \norm{\vect{A}_{\cdot j}}^2$.

The KL term can be expanded as:
\begin{equation}
    \frac12\sum_{j=1}^D\big(\mu_j^2+\sigma_j^2-\log(\sigma_j^2)\big),
\end{equation}
where $D$ is the dimension of the latent variable.

By introducing $\tilde{\vect{z}} := s\,\boldsymbol{\mu}$ and retaining only terms that depend on $\tilde{\vect{z}}$, $\boldsymbol{\sigma}$, or $s$, the loss function can be rewritten as
\begin{equation}
    \mathcal L(\tilde{\vect{z}},\boldsymbol{\sigma},s)
    =
    \frac{1}{2\sigma_{obs}^2}\norm{\vect{x}-\vect{A}\tilde{\vect{z}}}^2
    +
    \frac{s^2}{2\sigma_{obs}^2}\sum_{j=1}^D v_j\,\sigma_j^2
    +
    \frac12\sum_{j=1}^D\Big(\frac{\tilde{z}_j^2}{s^2}+\sigma_j^2-\log(\sigma_j^2)\Big)
    +
    \frac12(\log s)^2.
\end{equation}

The ELBO is thus dependent on $\tilde{\vect{z}}$, $\boldsymbol{\sigma}$, and $s$. We can consider the ELBO as a means to find the variational posterior that best fits the true posterior. As the ELBO approximates the log marginal posterior, we can write:
\begin{equation}
    \log p(\vect{x}) = -\mathcal L(\tilde{\vect{z}},\boldsymbol{\sigma},s) + \mathrm{KL}\!\left(q(\vect{z}, s\mid \vect{x})\,\|\, p(\vect{z}, s\mid \vect{x})\right)
\end{equation}
Differentiating this equation according to any of the variational posterior parameters yields a left hand side of the equation being equal to 0, as it only depends on the generative model. Thus, the derivative of the ELBO is equal to the negative derivative of the KL component, meaning that maximization of the ELBO with respect to variational posterior parameters is equivalent to minimizing the KL between the variational posterior and the true posterior. 

Finding the stationary value of the ELBO with respect to the different parameters yields a range of valuable insights. 

Differentiating the ELBO with respect to $\sigma_j$ yields for each $j=1,\dots,D$
\begin{equation}
    \Big(1+\frac{s^2}{\sigma_{obs}^2}v_j\Big)\sigma_j^2=1,
    \qquad
    \sigma_j^2(s)=\frac{1}{1+\frac{s^2}{\sigma_{obs}^2}v_j}.
    \label{eq:elboextsigma}
\end{equation}
This indicates that EA-VAE ensures that for small s, the standard deviation of the posterior is equal to that of the prior and increasing $s$ results in decreasing $\sigma_j$. Note also that under the assumptions of this analysis, eliminating $s$ results in a constant posterior variance, confirming stimulus-independent uncertainty for a standard VAE. 

Differentiating the ELBO with respect to $\tilde{\vect{z}}$ yields
\begin{equation}
    \Big(\vect{A}^\top \vect{A}+\frac{\sigma_{obs}^2}{s^2}I\Big)\tilde{\vect{z}} = \vect{A}^\top \vect{x},
    \qquad
    \tilde{\vect{z}}(s)=\Big(\vect{A}^\top \vect{A}+\frac{\sigma_{obs}^2}{s^2}\vect{I}\Big)^{-1}\vect{A}^\top \vect{x}, 
    \label{eq:elboextz}
\end{equation}
while we obtain
\begin{equation}
    \frac{s^4}{\sigma_{obs}^2}\sum_{j=1}^D v_j\sigma_j^2
    -
    \norm{\tilde{\vect{z}}}^2 + s^2\log s =0.
    \label{eq:elboexts}
\end{equation}
for $s$.

When $s$ is large, i.e.~ the image is dominated by the latents instead of observation noise, Eqs.~(\ref{eq:elboextsigma})--(\ref{eq:elboextz}) can be approximated by 
\begin{eqnarray}
    \sigma_j^2(s)&=&\frac{\sigma_{obs}^2}{s^2 v_j}\\
    \tilde{\vect{z}}(s)&=&(\vect{A}^\top \vect{A})^{-1}\vect{A}^\top \vect{x} \label{eq:zproj}
\end{eqnarray}
Substituting these expressions into Eq.~(\ref{eq:elboexts}) yields
\begin{equation}
    D s^2 - \norm{(\vect{A}^\top \vect{A})^{-1}\vect{A}^\top \vect{x}}^2 + s^2 \log s = 0.
\end{equation}
Note that $\vect{A}^+=(\vect{A}^\top \vect{A})^{-1}\vect{A}^\top$ projects the image $\vect{x}$ to the latent space and $\vect{A}$ provides the inverse mapping from latent samples back to image space. Using this notation and rearranging the equation:
\begin{equation}
    s^2\big(D+\log s\big)=\norm{\vect{A}^+\vect{x}}^2.
    \label{eq:svsx}
\end{equation}

To understand how $s$ is related to stimulus properties, we dissect the generative model into elementary transformations through singular value decomposition. Using the decomposition $\vect{A}=\vect{U}_r\boldsymbol{\Sigma}_r \vect{V}_r^\top$, such that $\boldsymbol{\Sigma}_r=\diag(\rho_1,\dots,\rho_r)$ (where $r \leq D$), we can transform Eq.~(\ref{eq:svsx}):
\begin{equation}
    s^2\big(D+\log s\big)= \sum_{k=1}^r\frac{(\vect{u}_k^\top \vect{x})^2}{\rho_k^2},
    \label{eq:svsAplus}
\end{equation}
where $\vect{u}_k$ are column vectors of $\vect{U}_r$.

In order to see how $s$ depends on interpretable properties of the image, we introduce the reconstructed image:
\begin{equation}
    \hat{\vect{x}} = \vect{A}\tilde{\vect{z}} \;\approx\; \vect{A} \vect{A}^+ \vect{x} = \vect{U}_r \vect{U}_r^\top \vect{x}
\end{equation}
This is expression highlights that the reconstructed image is a projection of the original image onto the linear plane defined by the generative model. The deviation of the original image from this plane can be characterized through the cosine similarity: 
\begin{equation}
    \cos(\vect{x},\hat{\vect{x}})=\frac{\norm{\hat{\vect{x}}}}{\norm{\vect{x}}},
\end{equation}
Given these two expressions for the reconstructed image, $\hat{\vect{x}}$ we expand Eq.~(\ref{eq:svsAplus}) with $\norm{\vect{\hat{x}}}^2 / \norm{\vect{\hat{x}}}^2$, which yields:
\begin{eqnarray}    
    s^2\big(D+\log s\big) &=& \sum_{k=1}^r\frac{(\vect{u}_k^\top \vect{x})^2}{\rho_k^2} 
    \frac{
    {\norm{\vect{x}}}^2 \cos(\vect{x},\hat{\vect{x}})^2
    }{
    \sum_{\ell=1}^r (\vect{u}_\ell^\top{\vect{x}})^2
    }\nonumber\\
    & = & 
    {\norm{\vect{x}}}^2 \cos(\vect{x},\hat{\vect{x}})^2 \sum_{k=1}^r w_k\frac{1}{\rho_k^2},
\end{eqnarray}
where we have introduced the in-plane weights of the projection, $w_k:=\frac{(\vect{u}_k^\top \vect{x})^2}{\sum_{\ell=1}^r (\vect{u}_\ell^\top \vect{x})^2}$, which sums up to 1. Intuitively, this equation establishes a connection between the scaling variable, $s$, the stimulus contrast, ${\norm{\vect{x}}}$, and the OOD-property of the image $\cos(\vect{x},\hat{\vect{x}})$, highlighting that $s$ decreases not only with contrast but also when the image is deviating from the plane defined by the generative model. Taking this results together with Eq.~(\ref{eq:elboextsigma}), an image that is off from the generative manifolds results in increased posterior variance. 

\section{Supplementary Methods and Experiments}

\subsection{Cost function and implementation for VAE and EA-VAE variants}\label{subsec:implementation}
\subsubsection{Inference on natural images}

Given the selected parametrization for prior, posterior and likelihood described in the main text \Cref{subsec:method-vae-eavae-nat}, \Cref{eqn:VAEcost} takes the form of
%
\begin{align}
\mathcal{L}_{VAE}(\mathbf{x},\psi,\phi) =  \frac{1}{2\sigma_{obs}^2} \sum_{i=1}^{M}(\hat{\mathrm{x}}_i-\mathrm{x}_i)^2 
+ \frac{M}{2}\log{(2\pi\sigma_{obs}^2)} \nonumber
\\
 + \beta \sum_{j=1}^{D} \left(-1+|\mu_j(\mathbf{x})|-\log{(b_j(\mathbf{x})})
+b_j(\mathbf{x})\exp{(-\frac{|\mu_j(\mathbf{x})|}{b_j(\mathbf{x})})}\right).
\label{eqn:SVAEcost}
\end{align}

It is worth mentioning that in cases when the observation noise is assumed to be drawn from an uncorrelated Normal distribution, then \Cref{eqn:VAEcost} can be rewritten by global multiplication of $2 \sigma_{obs}^2$ so that  the $\beta$ hyperparameter absorbes the assumed observation noise \cite{rybkin2021}, that is
\begin{equation}
\beta '= 2 \sigma_{obs}^2 \beta.
\label{eqn:beta}
\end{equation}

\noindent Further normalization can as well turn the first term in \Cref{eqn:VAEcost} into the mean squared error (MSE). Selecting a larger $\beta$ as done in $\beta$-VAEs then corresponds to performing inference assuming an enlarged observation noise, fostering disentanglement by use of an uncorrelated prior.

\begin{figure*}
\centering
\includegraphics[width=1.0\linewidth]{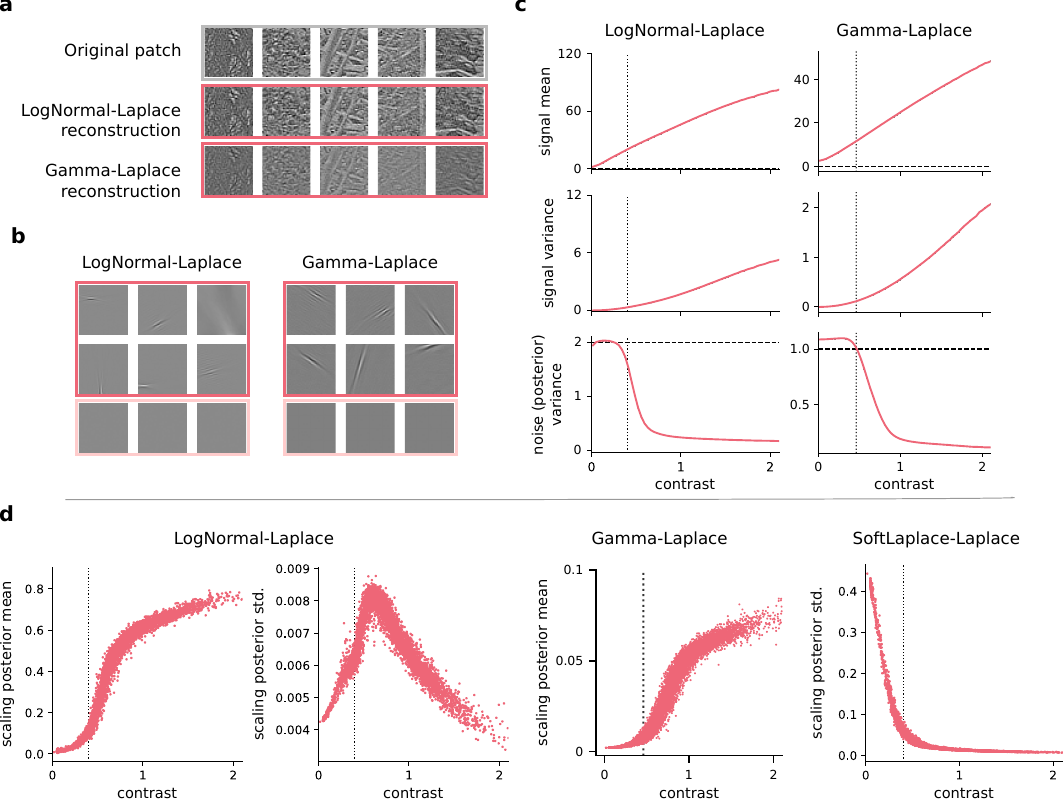}
\caption{\textbf{Properties of inferred posteriors as contrast is varied for EA-VAE models trained on natural image patches for variations on how scaling variable is modeled: either from a LogNormal, Gamma or Softplus-activated Laplace distribution.} \textbf{a}, Example test image patches and their respective reconstruction through EA-VAE. \textbf{b}, Receptive filters of example latents in the EA-VAEs. \textbf{c}, Signal mean, signal variance and noise variance of the inferred latent posteriors in natural image-trained EA-VAEs as a function of image contrast. Prior mean and variance are shown in \emph{horizontal dashed black lines}. Observation noise assumed in the model is shown in \emph{vertical dotted black lines}. \textbf{d}, Inferred posterior mean and posterior std. of the scaling variable for individual patches (dots) in the EA-VAE model, as a function of the measured contrast of these images. Observation noise assumed in the model is shown in \emph{vertical dotted black lines}.}
\label{SIfig:natural}
\hrulefill
\end{figure*}

In the case of the EA-VAE, when modeling the scaling variable as a softplus activated Laplace distribution, the loss function in \Cref{eqn:EA-VAEcost} takes the following form 
\begin{align}
\mathcal{L}_{EA-VAE}(\mathbf{x},\psi,\phi,\chi) =  
    \frac{1}{2\sigma_{obs}^2} \sum_{i=1}^{M}(\hat{\mathrm{x}}_i-\mathrm{x}_i)^2 
    + \frac{M}{2}\log{(2\pi\sigma_{obs}^2)}\nonumber 
\\
    + \beta_1 \sum_{j=1}^{D-1} \left(-1+|\mu_j(\mathbf{x})|-\log{(b_j(\mathbf{x})})+b_j(\mathbf{x})\exp{(-\frac{|\mu_j(\mathbf{x})|}{b_j(\mathbf{x})})}\right) \nonumber
\\
    + \beta_2 \left( 
        -1+|\mu_s(\mathbf{x})|-\log{(b_s(\mathbf{x}))}+
        b_s(\mathbf{x})\exp{(-\frac{|\mu_s(\mathbf{x})|}{b_s(\mathbf{x})})} \right),
\label{eqn:EA-SVAEcost}
\end{align}
\noindent where $\mu_s$ and $b_s$ are the inferred parameters of the scaling variable posterior. Importantly, to derive \Cref{eqn:EA-SVAEcost} it was taken into account that any invertible transformation applied to $s$ ensures that the Kullback-Leibler (KL) divergence in the cost function remains unaffected\cite{gil2013renyi}. We note that the resulting distribution (Laplace transformed by Softplus activation) does not have tractable moments. Thus we approximated the true posterior mean with samples, given that the inferred variance of $s$ was low except below the observation noise, where slightly different samples become similarly close to 0 due to softplus activation (\Cref{SIfig:natural}). Also note the substitution of $D$ in the first term of \Cref{eqn:SVAEcost} for $D-1$ in \Cref{eqn:EA-SVAEcost}, since we keep the overall number of latent variables unchanged for a fair comparison.

When modeling the scaling variable as a Log-normal distribution, we trained a version of the EA-VAE with a $\textrm{Log-normal}(0,1)$ prior for $s$. Similar to the homogeneous encoder, we use a joint encoder. This encoder produces mean and scale for Laplace variational posterior and a mean and variance of a Normal for the lognormal scaling variable, which is exponentiated after sampling from a Normal distribution (see \Cref{fig:archs}a for detailed architecture). Same as the main text model, $\beta_2$ was decreased linearly from 10 to 1 in the course of the first 100 epochs. Here, the loss function of the model resembles \Cref{eqn:EA-SVAEcost}, with the second KL term replaced with the respective KL for Normal distributions
\begin{align}
\mathcal{L}_{EA-VAE}(\mathbf{x},\psi,\phi,\chi) =  
    \frac{1}{2\sigma_{obs}^2} \sum_{i=1}^{M}(\hat{\mathrm{x}}_i-\mathrm{x}_i)^2 
    + \frac{M}{2}\log{(2\pi\sigma_{obs}^2)}\nonumber 
\\
    + \beta_1 \sum_{j=1}^{D-1} \left(-1+|\mu_j(\mathbf{x})|-\log{(b_j(\mathbf{x}))}+b_j(\mathbf{x})\exp{(-\frac{|\mu_j(\mathbf{x})|}{b_j(\mathbf{x})})}\right) \nonumber
\\
    + \beta_2 \frac{1}{2}\left( 
        -1+(\nu(\mathbf{x}))^2-\log{(\xi^2(\mathbf{x}))}+
        \xi^2(\mathbf{x})\right),
\label{eqn:EA-SVAEcost-lognormal}
\end{align}
\noindent where $\nu$ and $\xi^2$ are the mean and variance of a Normal distribution for the lognormal scaling variable.

The third variant modeled the scaling variable as a Gamma distribution, which naturally restricts the scaling variable to positive values. The Gamma distribution can be parametrized in terms of the shape and scale parameters $k$ and $\theta$ respectively. For simplicity, we assumed that posteriors $q_\chi(s|\mathbf{x})=\textrm{Gamma}(s;k,\theta(\mathbf{x}))$ have a fixed shape parameter $k=2$.
Since $k$ is fixed, the posterior's mean, mode and variance depend only on the inferred scale parameter $\theta$. For the prior distribution, the shape parameter was also fixed at $k=2$ and the scale parameter $\theta=1/\sqrt{2}$ was chosen such that the prior had unit variance $k\theta^2=1$. A constant shape parameter $k$ means that the posterior standard deviation is always proportional to the posterior mean. This choice is in line with Weber's Law of perception \cite{weber1834pulsu}, where the estimation error of a quantity is a constant fraction of its mean (see \Cref{fig:archs}b for detailed architecture). 
In this case, \Cref{eqn:EA-VAEcost} can be expressed similar to \Cref{eqn:EA-SVAEcost}, with the second KL term based upon Gamma distributions instead of Laplace \cite{penny2001}

\begin{align}
\mathcal{L}_{EA-VAE}(\mathbf{x},\psi,\phi,\chi) =  
    \frac{1}{2\sigma_{obs}^2} \sum_{i=1}^{M}(\hat{\mathrm{x}}_i-\mathrm{x}_i)^2 \nonumber
\\
    + \beta_1 \sum_{j=1}^{D-1} \left(-1+|\mu_j(\mathbf{x})|-\log{(b_j(\mathbf{x}))}+b_j(\mathbf{x})\exp{(-\frac{|\mu_j(\mathbf{x})|}{b_j(\mathbf{x})})}\right) \nonumber
\\
    + \beta_2 \, 2(-\log\sqrt{2}-\log\theta(\mathbf{x})+\sqrt{2}\,\theta(\mathbf{x})-1).
\label{eqn:EA-SVAEcost-gamma}
\end{align}
Note as well that because of the normalization done to the Van Hateren dateset in this case, observation noise $\sigma_{obs}$ gets rescaled and reduced, which we absorb into the $\beta_1'$ and $\beta_2'$ hyperparameters (Supplementary Table S1).
\subsubsection{Inference in other classic computer vision domains}
The cost functions constructed for the MNIST, ChestMNIST and NIH-ChestXray14 models under the assumptions described in the main text \Cref{subsec:method-vaeeavae-nist}, were
\begin{align}
\mathcal{L}_{VAE}(\mathbf{x},\psi,\phi) =  \sum_{i=1}^{M}-(\mathrm{x}_i \log(\hat{\mathrm{x}}_i)+(1-\mathrm{x}_i)\log(1-\hat{\mathrm{x}}_i)) \nonumber\\
+\beta' \sum_{j=1}^{D} \frac{1}{2}\left(-1+(\mu_j(\mathbf{x}))^2-\log{(\sigma_j^2(\mathbf{x}))}
+\sigma_j^2(\mathbf{x})\right),
\end{align}
\begin{align}
\mathcal{L}_{EA-VAE}(\mathbf{x},\psi,\phi,\chi) =  \sum_{i=1}^{M}-(\mathrm{x}_i \log(\hat{\mathrm{x}}_i)+(1-\mathrm{x}_i)\log(1-\hat{\mathrm{x}}_i)) \nonumber\\
+\beta'_1 \sum_{j=1}^{D-1} \frac{1}{2}\left(-1+(\mu_j(\mathbf{x}))^2-\log{(\sigma_j^2(\mathbf{x}))}
+\sigma_j^2(\mathbf{x})\right) \nonumber\\
+\beta'_2 \frac{1}{2}\left(-1+(\nu(\mathbf{x}))^2-\log{(\xi^2(\mathbf{x})}
+\xi^2(\mathbf{x})\right),
\label{eqn:VAE-EAVAE-costmnist}
\end{align}
\noindent while for the cMNIST, cFashionMNIST and cChestMNIST, the first term was correspondingly replaced by the sum of squared errors loss, that is $\sum_{i=1}^{M}(\hat{\mathrm{x}}_i-\mathrm{x}_i)^2$.

\subsection{Post-training evaluation of VAEs and EA-VAEs for natural images}\label{subsec:evaluation-nat}

\subsubsection{Latent receptive fields} \label{subsec:receptive_fields}
For fully connected models, as employed for the natural image dataset, the receptive field $\mathbf{RF}_j$ associated to the $j$-th latent dimension was computed through the Spike Triggered Average (STA) technique on the test dataset for all $j=1,...,1800$. The STA was estimated as

\begin{equation}
\mathbf{RF}_j =  \underset{\left \{ \mathbf{x} \right \} }{\textrm{mean}}\,\mu_j(\mathbf{x})\,\mathbf{x},
\end{equation}

\noindent where $\mu_j(\mathbf{x})$ is the $j$-th component of the mean of $\mathbf{x}$'s inferred posterior $q_\phi(\mathbf{z}|\mathbf{x})$.

Latent units were discerned as informative or non-informative according to their contribution to the dataset reconstruction. When computing the relative change in the images reconstructed when all units are active, or when a specific unit is replaced by its mean activation throughout the ensemble, it was observed that the units associated to orientation-sensitive filters are clustered with the most contribution, while the second group barely contribute.
All results shown in the main text were computed using only the informative latent dimensions.

\subsubsection{Uncertainty representation in the latent space}\label{subsec:uncertainty_latent}

We quantified the response of neurons for different observations and levels of uncertainty through the signal mean, signal variance and noise variance as a function of contrast.

For this, given an ensemble of patches $\left \{ \mathbf{x} \right \}$, we infer the contrast $s^*(\mathbf{x})$ for each patch as the across-pixel standard deviation of intensities: $s^*(\mathbf{x}) = \sigma_{pix}(\mathbf{x})$.

The signal mean is defined as the average distance of the posterior's mean to the prior's mean (0 in this case) over all patches that have the same inferred contrast.
Signal variance is computed as the variance of the posterior's mean of patches of same contrast, averaged across latent dimensions.
Finally, noise variance is defined as the posterior's variance vector $\boldsymbol{\sigma}^2(\mathbf{x})$ averaged across latent dimensions and observations of equal inferred contrast $s$.

These quantities were computed for the test set, where no two patches have the same exact contrast. Hence, these were binarized with small enough bins such that no jumps were visible in the plots, and with error bars comparable to line-widths. The binarized expressions are

\begin{align}
\text{SM}(s_i) = \underset{\left \{ \mathbf{x} | s^*(\mathbf{x})\in B_i \right \}}{\textrm{mean}}  \left \| \boldsymbol{\mu}(\mathbf{x}) \right \|\\
\text{SV}(s_i) = \underset{j}{\textrm{mean}}\underset{\left \{ \mathbf{x} | s^*(\mathbf{x})\in B_i \right \}}{\textrm{Var}}  \mu_j(\mathbf{x}) \\
\text{NV}(s_i) = \underset{j}{\textrm{mean}}\underset{\left \{ \mathbf{x} | s^*(\mathbf{x})\in B_i \right \}}{\textrm{mean}}  \sigma_j^2(\mathbf{x})
\end{align}

\noindent The associated errors can be expressed as

\begin{align}
\epsilon_{_{\text{SM}}}(s_i)=\frac{1}{\sqrt{N_i}}\sqrt{\underset{\left\{\mathbf{x}|s^*(\mathbf{x})\in B_i\right\}}{\textrm{Var}} \left \| \boldsymbol{\mu}(\mathbf{x}) \right \| } \\
\epsilon_{_{\text{SV}}}(s_i)=\frac{1}{\sqrt{D}}\underset{j}{\textrm{mean}} \sqrt{\frac{2}{(N_i-1)}}\underset{\left\{\mathbf{x}|s^*(\mathbf{x})\in B_i\right\}}{\textrm{Var}}  \mu_j(\mathbf{x}) \\
\epsilon_{_{\text{NV}}}(s_i)=\frac{1}{\sqrt{N_i}\sqrt{D}}\sqrt{\underset{\left\{j\cup \mathbf{x}|s^*(\mathbf{x})\in B_i\right\}}{\textrm{Var}}\boldsymbol{\sigma}^2(\mathbf{x})}.
\end{align}

It is worth mentioning that the natural scenes dataset used in this work has patches with varied values of pixel intensity standard deviation $\sigma_{pix}(\mathbf{x})$. As patches with high contrast live far away from the origin in the M-dimensional pixel space (since $s_{pix}(\mathbf{x}) = \sqrt{\frac{1}{M-1}\sum_{i=1}^{M}(x_i-\bar x)^2} = \sqrt{\frac{1}{M-1}}\left \| \mathbf{x} \right \| $), it is expected that the network will have more difficulty exploring farther regions in the pixel space than the regions where low-contrast patches live. Therefore, in evaluation instances, patches with contrast higher than $\sigma_{pix}=2$ were not considered.

\subsubsection{Divisive normalization experiments} \label{subsec:div_norm}

To characterize the dependencies between latents of the Variational Autoencoders, we calculated conditional distributions for pairs of latents by analyzing the distribution of posterior means for a large number of natural image patches. Across a large number of randomly sampled filter pairs the variance of the conditional distribution was calculated at four quadrants: two central and two flanking quadrants (\Cref{fig:dn}a), which corresponded to low-level, and high-level activations of the conditioned filter, respectively (\Cref{fig:dn}b). 

We modeled divisive normalization by predicting responses of individual latents of the Variational Autoencoder through the linear filter response of that latent divided by the weighted sum of the (linear) responses of all other latents
\begin{equation}
    z_{i}=\frac{L_{i}}{\sqrt{\sum_{j\neq i}w_{ji}L_{j}^{2} + \sigma^{2}}}.
    \label{eq:dn}
\end{equation}
\noindent
Here $L_j$ is the linear response of latent variable $j$ and weights are constrained to be positive. Weights, $w_{ji}$ were determined through regression. Linear responses of latents were calculated by taking the dot product of the image and the vector mapping the latent to the pixel space in the generative model. Fitting of the model was driven by the deviation of the linear responses from the posterior produced by the recognition model. 

\subsection{Post-training evaluation of VAEs and EA-VAEs for MNIST, ChestMNIST and NIH-ChestXray14}\label{subsec:evaluation-nist}

\subsubsection{Out of distribution experiments}\label{SM_subsubsec:ood}
\begin{figure*}
\centering
\includegraphics[width=1.0\linewidth]{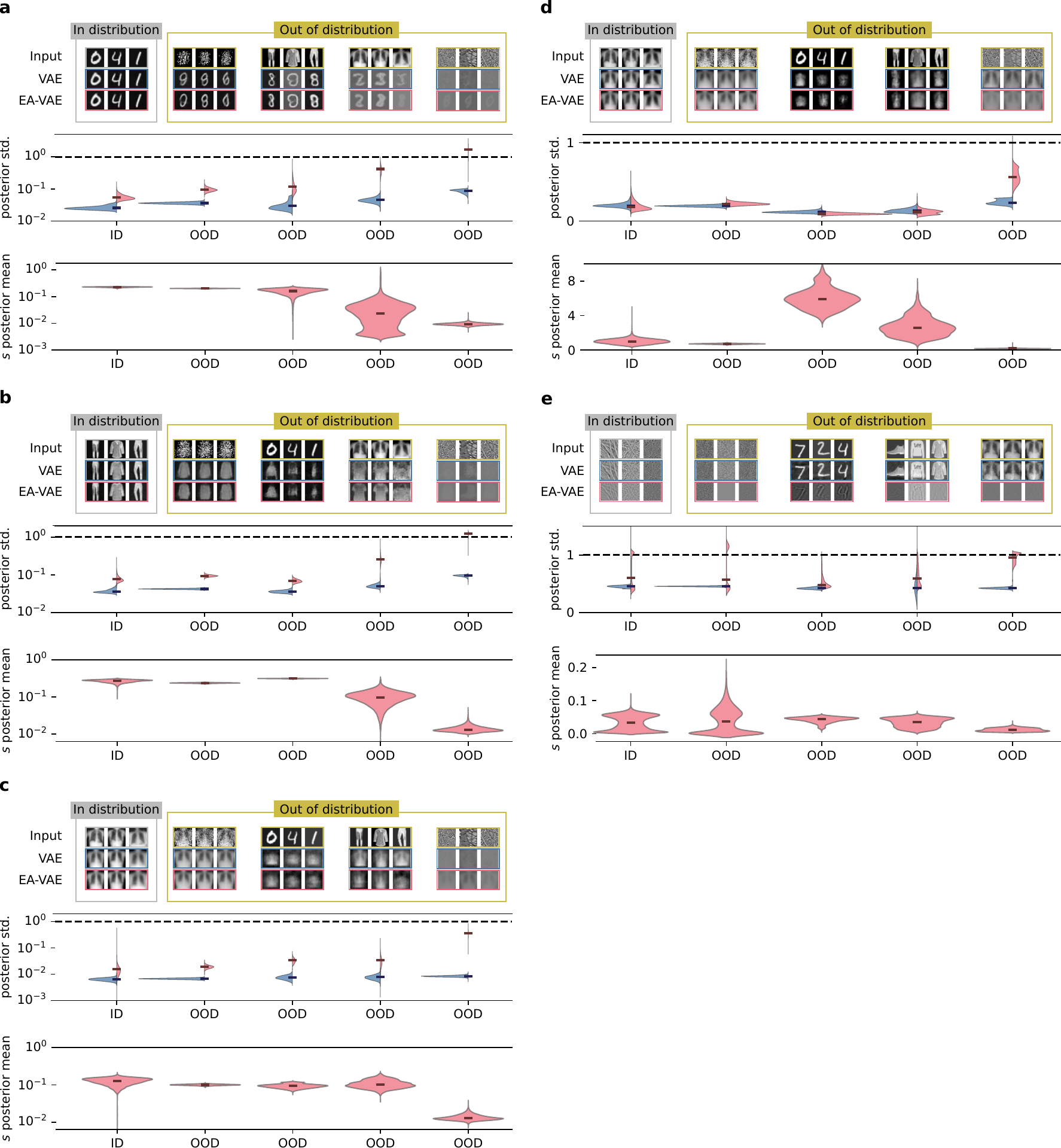}
\caption{\textbf{Out of distribution experiments.} Same as in Fig. 3a, now when training on \textbf{a}, cMNIST 10D, \textbf{b}, cFashionMNIST 10D, \textbf{c}, cChestMNIST, \textbf{d},  ChestMNIST and \textbf{e}, Gamma-Laplace Van Hateren ($z$-scored).}
\label{SIfig:ood}
\hrulefill
\end{figure*}

To study the capacity of the trained models to associate near-to-prior uncertainty for images from previously unseen distributions, we compared the latent representations of images in and out of the distribution used for training. The reconstruction quality was analysed by projecting the posterior mean back to the pixel-space using the decoder to observe how OOD inputs were represented. Latent uncertainty to a given observation was computed by \Cref{eq:unc}.
We looked at models trained with all datasets (MNIST, ChestMNIST, cMNIST, cFashionMNIST, cChestMNIST and Van Hateren) and tested with the complementary domains and corresponding pixel-shuffled dataset (\Cref{fig:results-ood}a and \Cref{SIfig:ood}).
In all cases, images from the OOD datasets were resized to coincide the size of images from the training set, and rescaled in pixel intensity to match the in-distribution pixel intensity distribution. For the models trained on Van Hateren, that is population wise z-scoring the OOD datasets. For all other models trained on MNIST domains, that is rescaling the Van Hateren's dataset pixel intensity distribution such that it's mean $\pm$ 3 deviations fit the range [0,1].

The same findings as the ones described in the main text \Cref{subsec:main_ood} were replicated when training models with ChestMNIST and testing on natural images (\Cref{SIfig:ood}d, rightmost block), yet when testing on digits and fashion images, both models seem to fare comparably, with reported uncertainties that do not approach the prior, although the reconstructed images do not collapse to within domain examples. When training with natural images and testing on the other (\Cref{SIfig:ood}e), the distribution of reported uncertainties, which originally had two modes for within distribution images corresponding to more informative (high contrast) and less informative (low contrast) images, reduces to a single mode mode for VAEs and EA-VAEs. The asymmetry in the results when original and target domains are inverted suggests a hierarchy in the richness of the feature space learned from the statistics of the images.

\subsubsection{Image corruption}\label{subsec:image_corruption}
The image corruption experiment was done over the complete test set.
Each image was synthetically corrupted by applying a two-dimensional Gaussian kernel of covariance $\eta_b I_{2x2}$. The resulting blurred image was encoded into the models' latent space, where the measure of uncertainty was computed. We compute the uncertainty of the network for a given input by \Cref{eq:unc}. This process was repeated for all images, and the average uncertainty was computed. This was repeated for different $\eta_b$ values until the images were completely blurred out ($\eta_b=100$ in the case of NIH-ChestXray14 dataset). We repeated the experiment with pixel-noise corruption. In this case, each pixel was intensity corrupted by additive white Gaussian noise of null mean and variance $\eta_p^2$. The whole perturbed image was then rescaled in intensity to preserve the original range of intensities. $\eta_p$ values between 0 and 2.5 were considered, going from no perturbation to a strongly perturbed image.

\subsubsection{Uninformative observations}\label{subsec:uninformative}
\begin{figure*}
\centering
\includegraphics[width=0.5\linewidth]{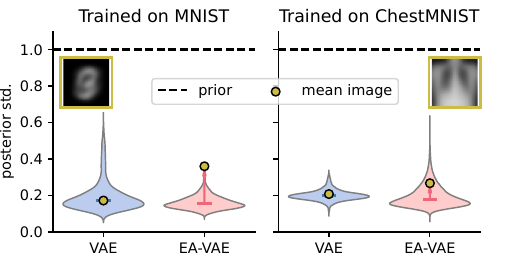}
\caption{Latent posterior width (\emph{mustard dots}, with color matching that of the frame of average images) for uninformative images (\emph{top}, \emph{mustard} framed images) in the MNIST (\emph{left}) and ChestMNIST (\emph{right}) data sets. The distribution of noise posterior widths (std.) for in-distribution test images for standard VAE and EA-VAE are presented as violin plots. Prior uncertainty is shown as a reference (dashed line).}
\label{SIfig:uninf-mnist}
\hrulefill
\end{figure*}

We evaluated the models' ability to handle uninformative observations by creating an ``average image'' through pixel-wise averaging of all dataset images. This image served as a proxy for an uninformative observation. As expected, this image resembles a blurry, unrecognizable digit (MNIST) or a slightly blurry x-ray (ChestMNIST) (\Cref{SIfig:uninf-mnist}). We characterize the latent uncertainty in the models measuring the noise standard deviation of the posterior and averaging along the $\mathbf{z}$ latent dimensions (\Cref{eq:unc}). In the case of MNIST, the VAE produced a latent uncertainty ($u_{_{\text{VAE}}}=0.17$) comparable to that of real handwritten digits ($u_{_{\text{ID}}}$) in the dataset [$p(u_{_{\text{ID}}}>u_{_{\text{VAE}}})=0.49$], and considerably below the unit standard deviation of the prior (\Cref{SIfig:uninf-mnist} left panel). Notably, the same measurement in the EA-VAE yielded $u_{_{\text{EA-VAE}}}=0.36$, This uncertainty is not only twice of that obtained for the VAE but also significantly higher than the uncertainty of actual samples from the distribution [$p(u_{_{\text{ID}}}>u_{_{\text{EA-VAE}}})= 10^{-4}$]. A tendency similar to that observed on the MNIST dataset can also be seen on ChestMNIST-trained VAE and EA-VAE models [$p(u_{_{\text{ID}}}>u_{_{\text{VAE}}})=0.33$; $p(u_{_{\text{ID}}}>u_{_{\text{EA-VAE}}}) = 0.12$] (\Cref{SIfig:uninf-mnist} right panel). Note, that the average x-ray image resembles actual data samples more closely than the average digit image, which explains the less pronounced advantage of EA-VAE over the standard VAE. In summary, these results show that uncertainty estimates in the EA-VAE, but not in the standard VAE, distinguish uninformative images from actual images. 

\subsubsection{Morphing and uncertainty}\label{subsec:morphing}
\begin{figure*}
\centering
\includegraphics[width=0.9\linewidth]{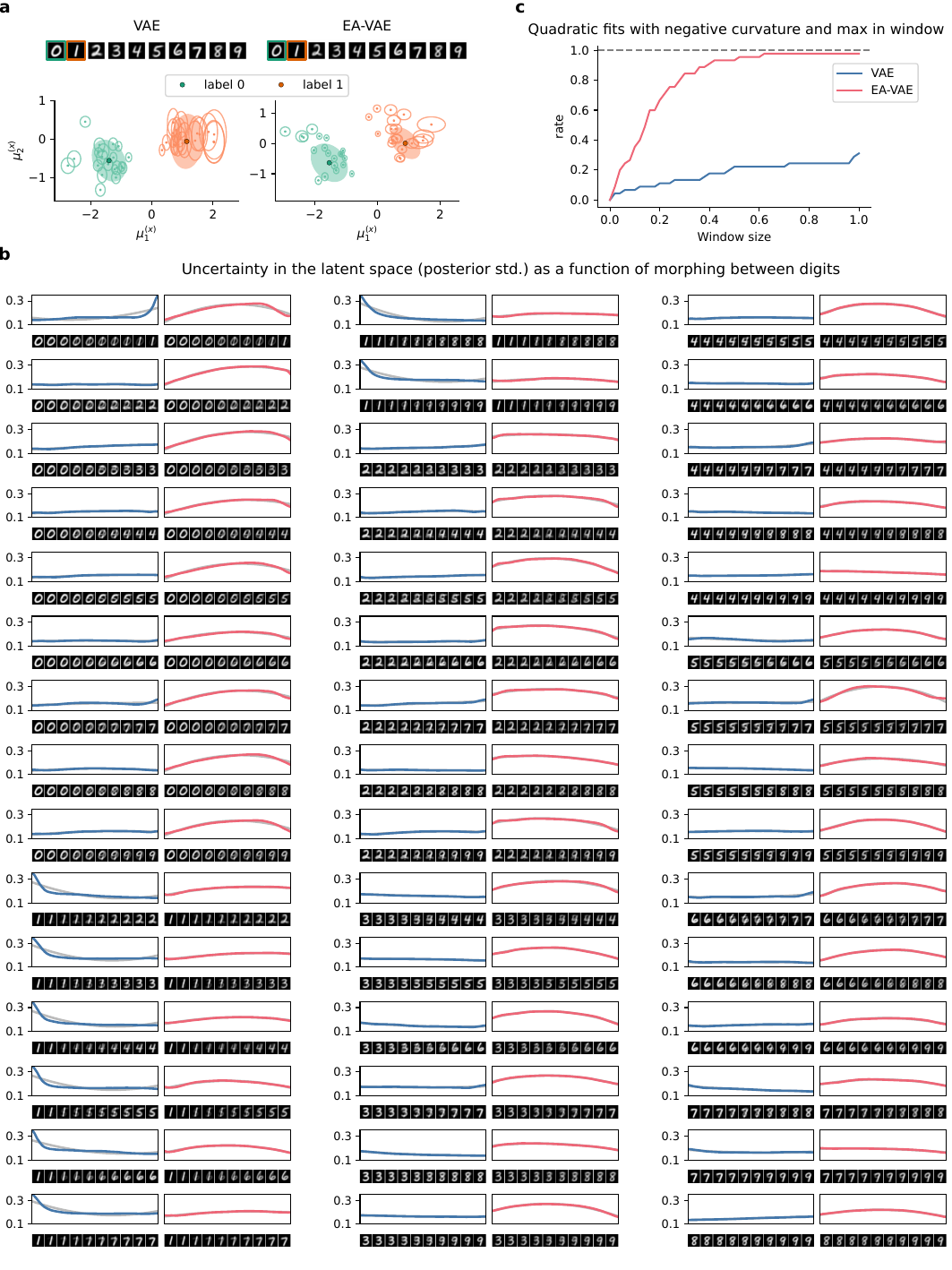}
\caption{\textbf{Morphing experiments in MNIST.} \textbf{a}, Centers for the distribution corresponding to each digit are first found. \textbf{b}, Posterior widths for the morphed digits with
standard VAE (left) and EA-VAE (right). Position where uncertainty is maximal is computed via quadratic fit (shown in gray curve). \textbf{c}, For a given window size around the middle point between the centers, the fraction of interpolations where the maximal  uncertainty lies within this window is computed via \textbf{b} fittings.}
\label{SIfig:morphing}
\hrulefill
\end{figure*}

We conducted an experiment to study the uncertainty in the VAE and EA-VAE latent space when morphing digits from one label to another.
The first step for doing this was selecting all the test images that belong to a specific label \textit{A} and computing the average coordinate in the latent space of these (\Cref{SIfig:morphing}a bottom panels). The output $\hat{\mathbf{x}}_A$ of the decoder from that position is a representative image of the label \textit{A} (\Cref{SIfig:morphing}a top). The same procedure is repeated with images belonging to label \textit{B} obtaining $\hat{\mathbf{x}}_B$. Then, morphing is done by linearly combining $\hat{\mathbf{x}}_A$ and $\hat{\mathbf{x}}_B$ with weight $\lambda$ varying from 0 to 1, that is

\begin{equation}
\hat{\mathbf{x}}_{AB,\lambda}=(1-\lambda) \hat{\mathbf{x}}_{A} + \lambda \hat{\mathbf{x}}_{B}.
\end{equation}

We then inferred the corresponding posteriors taking these morphed images as input, and computed the uncertainty of the network by \Cref{eq:unc}.

Through this procedure, uncertainty curves were computed for all the combinations of morphing from labels $0$ to $9$ (\Cref{SIfig:morphing}b color curves).

To evaluate if a certain curve has the expected behavior of being maximally uncertain at intermediate $\lambda$ values, we fitted a second order polynomial function (\Cref{SIfig:morphing}c gray curves) and checked whether it's curvature was negative and the maximum was located inside a window of size $L$ centered respect to $\lambda=0.5$, for different $L$ values (\Cref{SIfig:morphing}c). 

\subsubsection{Predictive uncertainty}\label{subsec:predictive_uncertainty}
To evaluate the role of uncertainty representations in classification, two separate MLPs were trained to classify MNIST handwritten digits into labels using samples from the posterior latent representations of VAEs and EA-VAEs. Once trained, the classification task involved two experiments. In the first, a set of continuously morphed digits was used to analyze predictive uncertainty, measured as the entropy of output distributions. We compared the entropy curves of the MLPs’ predictions by summing the entropy for all intermediate images for the VAE versus the EA-VAE. A paired t-test quantified the differences. In the second experiment, we tested the MLPs with out-of-distribution inputs, including ChestMNIST and pixel-shuffled MNIST images, and compared the entropy of predictions based on the two latent representations.

\subsubsection{MNIST MLP classifier}\label{subsec:mlp_classifier}
A multilayer perceptron (MLP) was trained to classify MNIST digits into its labels. Given a digit $\mathbf{x}$, the network receives as an input a sample $\mathbf{z}$ of the posterior distribution $q_\phi(\mathbf{z}|\mathbf{x})$ encoded by the already trained and fixed VAE (or EA-VAE), and outputs a 10-dimensional vector quantifying the predicted probability for each label $l$: $\mathbf{w}=\left[P(l=0),P(l=1),...,P(l=9) \right]$. The MLP had two hidden layers of 50 units with relu non-linearity. For the output layer, softmax activation function was used. The model was trained for 1000 epochs, where for each epoch, resampling of the posterior distributions was done for computing the input of the network. Adam optimizer was used.

\subsection{Uncertainty Representations for ImageNet in Inception Feature Space}\label{subsec:inception}

Inception-v3 \cite{szegedy2016rethinking} has become a canonical architecture for computer vision. The activations of this model, when pretrained with Imagenet \cite{deng2009imagenet, russakovsky2015imagenet}, are nowadays often employed to construct a representational space where to compare distributions of images: for instance between real and synthetically generated data \cite{heusel2017gans}.

\begin{figure*}
\centering
\includegraphics[width=1.0\linewidth]{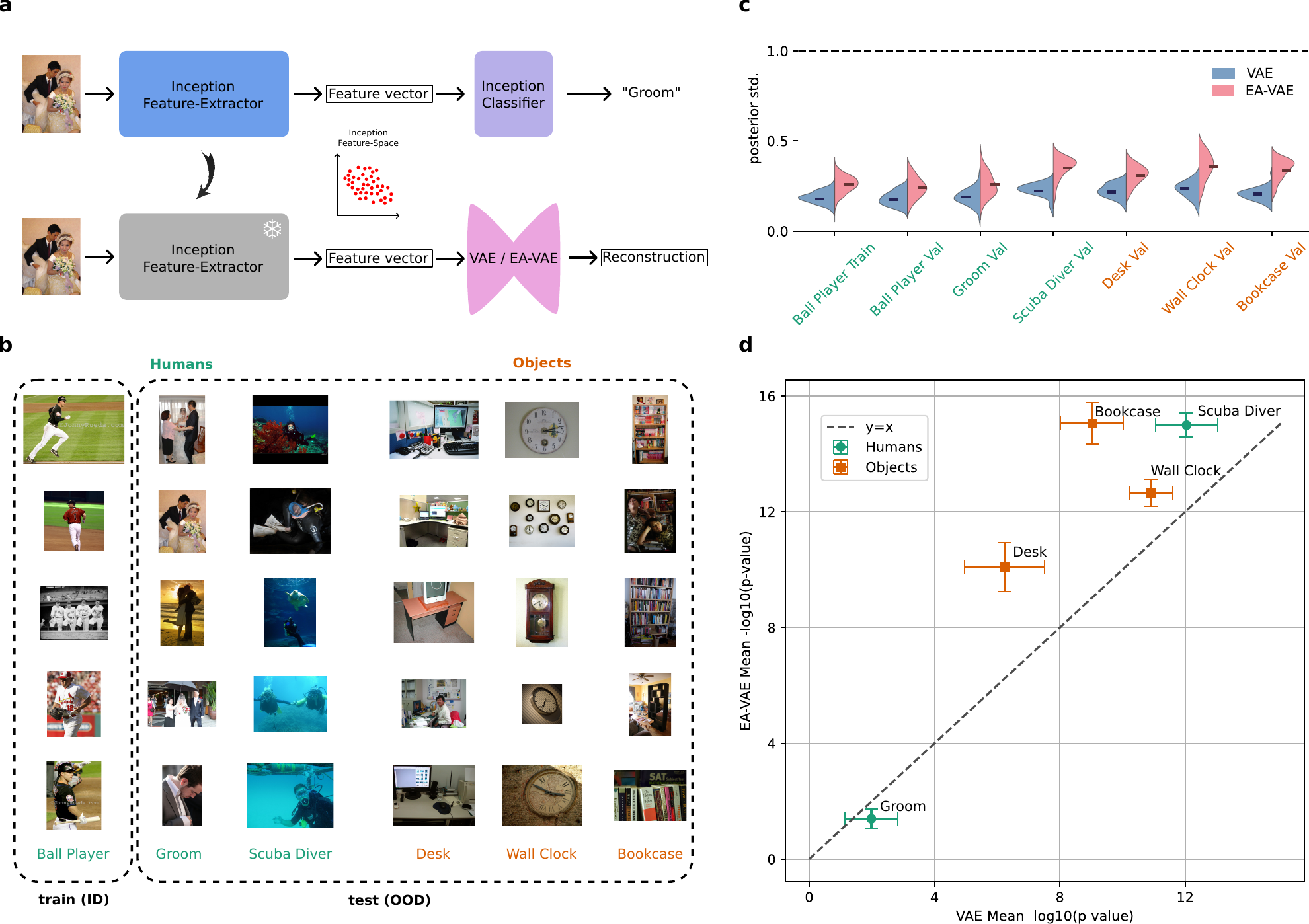}
\caption{\textbf{a}, VAE and EA-VAE trained on Imagenet from Inception feature space. \emph{Top}: The pretrained Inception network extracts for a given image, a feature vector (\emph{blue box}) from which the image is then classified (\emph{lilac box}). \emph{Bottom}: VAE/EA-VAE is trained using as input the feature vector extracted from pre-trained fixed Inception block. \textbf{b}, Example Imagenet images used to train and test the VAE/EA-VAE. \emph{Left}: Human-class (green label) Ball Player images were used to train. \emph{Right}: Human-class Groom and Scuba Diver, as well as Object-class (orange label) Desk, Wall Clock and Bookcase were used to test out of distribution examples. \textbf{c}, Posterior width for in-distribution (Ball Player) and out-of-distribution (Groom, Scuba Diver, Desk, Wall Clock and Bookcase) examples when trained on Inception feature vector for Ball Player images both for standard VAE (\emph{blue}) and EA-VAE (\emph{red}). As a reference, a dashed black line indicates the prior uncertainty. \textbf{d}, VAE vs EA-VAE capability to differentiate out-of-distribution examples. Comparison of posterior width for in-distribution and out-of-distribution for a single model is measured as the $\log_{10}(\text{p-value})$ for the t-test between the distributions. Each point corresponds to one of the out-of-distribution classes, and error bars arise from the $N=10$ trained models. As a reference, a dashed black line indicates the identity.}
\label{fig:inception}
\hrulefill
\end{figure*}

In this section, we compare the capabilities of VAEs and EA-VAEs to develop a meaningful representation of uncertainty, and detect distributional shifts in the Inception representational space. To that end, we first precompute Inception-v3 features for images from ImageNet (\Cref{fig:inception}a). We employ the three categories of human images present in Imagenet: `Ball Player', `Groom' and `Scuba Diver', and select another three categories: `Desk', `Wall Clock' and `Bookcase' (\Cref{fig:inception}b). To analyze whether VAEs and EA-VAEs can learn meaningful representations of uncertainty that capture the semantic similarity between classes, we train these models on the first of the human classes (`Ball Player'), and evaluate reported uncertainty on examples from the other human and object classes (\Cref{fig:inception}c). One would expect the models to be able to detect out of distribution examples, with higher uncertainty for those classes which are more dissimilar to the training set.

To systematically compare VAE and EA-VAE uncertainty representations for out of distribution images, we trained $N=10$ VAEs and EA-VAEs. Encoders and decoders are symmetric with two fully connected layers each (of 512 and 256 neurons with relu activation), representing posteriors with 64 latent dimensions. We kept the VAE and EA-VAE architectures as similar as possible to each other. The only difference being that in the EA-VAE the last of the 64 latents went through a soft-plus function before multiplying the output, instead of integrating the standard decoder as the rest of the variables. Otherwise, the cost function remains intact, and all latent variables have identical normal priors. For each model, we obtained the distribution of posterior latent uncertainties for each image class (\Cref{fig:inception}c), and then we computed the p-values for a t-test between the distribution of posterior uncertainties for out-of-distribution examples vs in-distribution examples for both models (\Cref{fig:inception}d). We observe that the EA-VAE shows consistently higher significance than the VAE in the t-test for object images. This tendency is reversed for the `Groom' category, indicating that the EA-VAE perceives these two human classes as more similar than the VAE. Initially one may have expected the `Scuba Diver' class to be regarded as more similar by these models, since the contain humans. We observe however that both models report very high uncertainty in this case, surpassing that of objects. Close inspection of the images (see third column in \Cref{fig:inception}b) shows that these images are indeed quite dissimilar to the rest, with underwater scenes, and a major shift in the distribution of colors and shapes.

\subsection{Uncertainty Representations for Importance Weighted Autoencoder}

Importance Weighted Autoencoders (IWAEs) are a flexible extension of the VAE family, enabling a richer latent representation by taking multiple samples from the posterior \cite{burda2015importance}. This family of models optimizes the standard ELBO while permitting a more complex implicit posterior distribution that loosens the usual independence assumptions on the latent variables \cite{cremer2017reinterpreting}. IWAEs could be principled candidates for modeling natural images, where latents have been shown to display strong nonlinear dependencies \cite{wainwright1999visual}.

IWAEs are trained by optimizing for the following objective, which is equivalent to the VAE ELBO in expectation:
\begin{equation}
    \mathcal{L}_{\mathrm{IWAE_k}}(\vect{x}, \psi, \phi) = 
    \mathbb{E}_{\vect{z_1}, \ldots, \vect{z_k} \sim q_\phi(\vect{z} \mid \vect{x})} 
    \left[ \log \left( \frac{1}{k} \sum_{i=1}^{k} \frac{p_\psi(\vect{x} \mid \vect{z_i}) p(\vect{z_i})}{q_\phi(\vect{z_i} \mid \vect{x})} \right) \right] 
    \label{eq:placeholder_label}
\end{equation}

Here, the encoder and decoder are parameterized by $\phi$ and $\psi$, respectively and $k$ is a fixed hyperparameter of the model. The output of the encoder, $q_\phi(\vect{z} \mid \vect{x})$ is the naive posterior obtained by the VAE independence assumptions. Drawing $k$ samples from the naive posterior and implementing an importance weighting over them leads to a richer posterior distribution. This does not take an explicit form, but it is possible to draw samples from this implicit distribution.

\begin{figure*}
\centering
\includegraphics[width=0.95\linewidth]{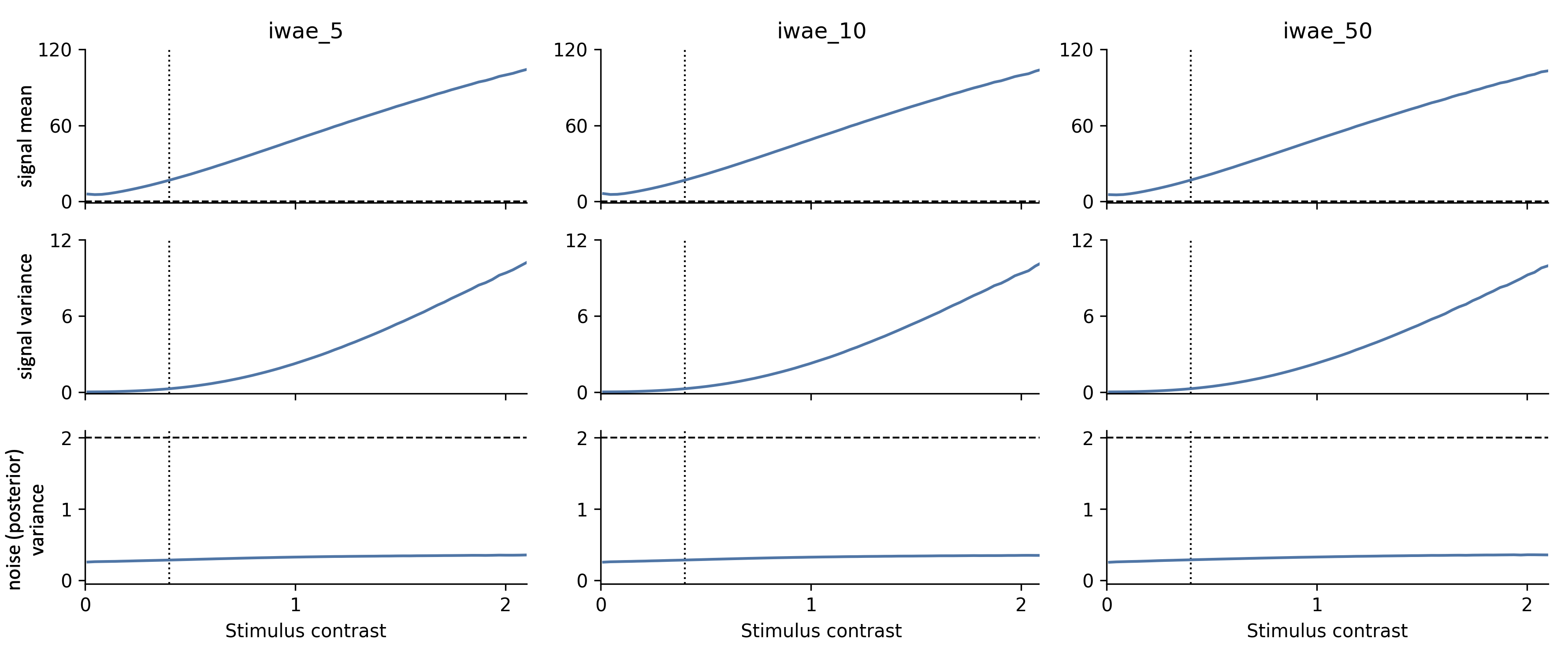}
\caption{Uncertainty estimates of the Importance Weighted Autoencoder on the Van Hateren dataset. The signal means, signal variances and noise (posterior) variances for the test images are shown as in the main text in three IWAE variants, with $k=5, 10, 50$ samples respectively. The updated architectures do not show any different representation than the standard VAE presented in the paper.}
\label{fig:iwae}
\hrulefill
\end{figure*}

We trained IWAEs on the Van Hateren dataset to see whether this extension leads to remedied uncertainty representation. The models were trained on 40x40 pixel patches with the same settings as the standard VAE. Three models were tested, using $k=5,10,50$ samples from the naive posterior distribution. For the analysis, we took 15 samples from the posterior and calculated the sample mean and sample variance for estimates of the implicit posterior moments. The results were no different than in the case of the standard VAE: posterior variance still fails to capture the uncertainty related to low-contrast images (\Cref{fig:iwae}). This shows that even a more flexible variant of the VAE is unable to perform a fundamental task of uncertainty estimation, which is remedied by the EA-VAE model.

\subsection{Uncertainty Representations for Monte Carlo Dropout VAE}

A common approach to approximate uncertainty in deep neural networks is Monte Carlo dropout \cite{gal2016dropout}, by maintaining dropout active during inference time and performing multiple stochastic forward passes\cite{kendall2017uncertainties}. To assess whether the EA-VAE achieves improved uncertainty representations beyond standard baselines, we compared it as well with a VAE with Monte Carlo dropout.

Following common practice in Bayesian deep learning, stochasticity (via dropout layers) was introduced exclusively in the encoder network, which models the approximate posterior over latent variables. The decoder was kept deterministic given the latent sample. During training, dropout was applied in the standard way, and during inference time, we kept dropout explicitly active.

At inference, for a given input image $\mathbf{x}$, the approximate posterior $\tilde{q}(\mathbf{z}|\mathbf{x})\approx\textrm{Normal}(\mathbf{z};\tilde{\boldsymbol{\mu}}(\mathbf{x}),\tilde{\boldsymbol{\sigma}}^2(\mathbf{x}))$ is then estimated through $T$ forward passes of the trained encoder, obtaining a set of $T$ latent posterior parameters from $q_\phi(\mathbf{z}|\mathbf{x})=\textrm{Normal}(\mathbf{z};\boldsymbol{\mu}(\mathbf{x}),\boldsymbol{\sigma}^2(\mathbf{x}))$
\begin{equation}
    \left\{ \boldsymbol{\mu}(\mathbf{x})_t,\boldsymbol{\sigma}^2(\mathbf{x})_t\right\}_{t=1}^T,
    \label{eq:Tparams}
\end{equation}
where each pair corresponds to one realization of the dropout mask.

The aggregated posterior mean $\tilde{\boldsymbol{\mu}}(\mathbf{x})$ and variance $\tilde{\boldsymbol{\sigma}}^2(\mathbf{x})$ for a single image $\mathbf{x}$ were then computed using the law of total expectation and law of total variance
\begin{equation}
\tilde{\boldsymbol{\mu}}(\mathbf{x}) = \mathbb{E}_t [\boldsymbol{\mu}(\mathbf{x})_t]
    \label{eq:MCtotalmean}
\end{equation}    
\begin{equation}
    \tilde{\boldsymbol{\sigma}}^2(\mathbf{x}) = \mathbb{E}_t [\boldsymbol{\sigma}^2(\mathbf{x})_t] + \text{Var}_t[\boldsymbol{\mu}(\mathbf{x})_t]
    \label{eq:MCtotalvar}
\end{equation}

The VAE with Monte Carlo dropout was trained on the manuscript digits MNIST dataset, following the same architecture and hyperparameters used for the MNIST-trained standard VAE and EA-VAE (\Cref{fig:archs}c and Table \Cref{tab:hyperparams-mnist}). The model was trained with dropout probability hyperparameter set as $p=0.2$ and $p=0.5$, without observing qualitative differences in the results.

\begin{figure*}
\centering
\includegraphics[width=0.95\linewidth]{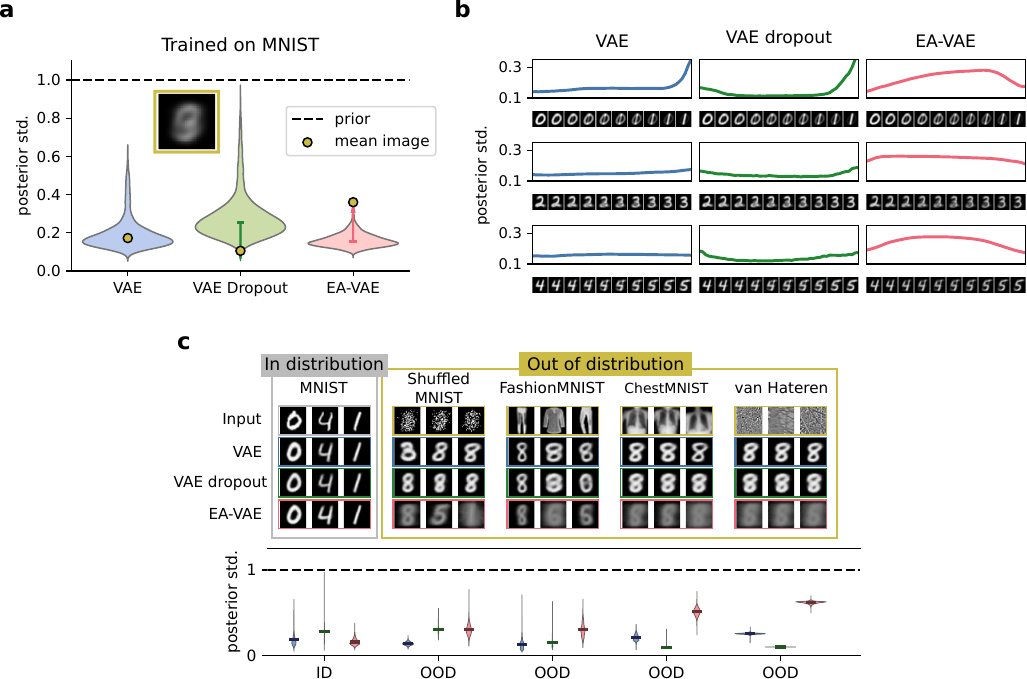}
\caption{\textbf{Characterization of posterior width for systematic image manipulations that affect inference uncertainty for VAE with active dropout during training in the encoder of the VAE, in comparison to standard VAE and EA-VAE.} \textbf{a}, Latent posterior width (\emph{mustard dots}, with color matching that of the frame of average image) for uninformative images (\emph{top}, \emph{mustard} framed images) in the MNIST dataset. The distribution of noise posterior widths (std.) for in-distribution test images for each model are presented as violin plots. Prior uncertainty is shown as a reference (dashed line). \textbf{b}, Posterior widths for gradual morphing between  representative image samples from digit 'i' and digit 'j' with standard VAE (\emph{left}), VAE with dropout (\emph{center}) and EA-VAE (\emph{right}). \textbf{c}, Posterior width for in-distribution (\emph{ID}) and out-of-distribution (\emph{OOD}) examples when trained on MNIST for standard VAE (\emph{blue}), VAE with dropout (\emph{green}) and EA-VAE (\emph{red}). \emph{Top}: Examples of images from the same domain (\emph{gray frames}) and different domains (\emph{mustard frames}) as inputs, and reconstructions by the standard VAE (\emph{second row}), VAE with dropout (\emph{third row}) and EA-VAE (\emph{fourth row}) respectively. \emph{Bottom}: uncertainty quantified by the posterior width for individual within-distribution or out-of-distribution images. As a reference, a dashed black line indicates the prior uncertainty.}
\label{fig:dropout}
\hrulefill
\end{figure*}

To evaluate the model in its capacity to represent uncertainty faithfully, we reproduced the experiments done for the standard VAE and EA-VAE on uninformative observations (main text \Cref{subsec:uninformative} and \Cref{fig:dropout}a), morphing digits (main text \Cref{subsec:main_morphing} and \Cref{fig:dropout}b) and out of distribution detection (main text \Cref{subsec:main_ood} and \Cref{fig:dropout}c). In all cases, the VAE with Monte Carlo dropout failed to represent uncertainty in a meaningful way, equivalently to how the Standard VAE behaved.

\clearpage

\section{Supplementary Tables and Figures}

\begin{table*}[ht]
\label{tab:hyperparams-natural}
\begin{center}
\caption{\textit{Hyperparameter settings for the Van Hateren models}}
\resizebox{\textwidth}{!}{%
\begin{tabular}{lccccccccccc}
\hline
Van Hateren Model & PC & $D$  & g( ) & $\sigma_{obs}$ & $\beta^{\text{VAE}}_1$ & $\beta_1^{\text{EA-VAE}}$ & $\beta_2^{\text{EA-VAE}}$ & lr & wd & Epochs & Batch Size \\ \hline
SoftLaplace-Laplace & 2 & 1800 & softplus & 0.4 & 0.01 $\rightarrow$ 1 & 1 & 10 $\rightarrow$ 1 & 3e-5 & 1e-6 & 5000 & 512 \\
LogNormal-Laplace & 2 & 1800 & exp & 0.4 & 0.01 $\rightarrow$ 1 & 1 & 10 $\rightarrow$ 1 & 3e-5 & 1e-6 & 5000 & 512 \\
Gamma-Laplace ($\alpha\approx5.315$) & 1 & 1800 & -- & 0.46/$\alpha\approx0.09$ & 1 & 1 & 2 & 3e-5 & 1e-5 & 10000 & 128 \\
\hline
\end{tabular}
}
\end{center}
\end{table*}

\begin{table*}[ht]
\label{tab:hyperparams-mnist}
\begin{center}
\caption{\textit{Hyperparameter settings for the NIST and NIH-ChestXray14 domain models}}
\resizebox{\textwidth}{!}{%
\begin{tabular}{lccccccccccc}
\hline
Model & PC & $D$ & h( ) & g( ) & $\sigma_{obs}$ & $\beta'_1$ & $\beta'_2$ & lr & wd & Epochs & Batch Size \\ \hline
MNIST & 1 & 5 & sigmoid & exp & -- & 4 & 1 & 1e-3 & 1e-5 &  500 & 128 \\
ChestMNIST & 1 & 5 & sigmoid & exp & -- & 1 & 1 & 1e-3 & 1e-5 &  500 & 128 \\
NIH-ChestXray14 & 1 & 20 & sigmoid & softplus & -- & 50 & 50 & 5e-4 & 1e-5 &  500 & 128 \\
cMNIST10D & 1 & 10 & -- & exp & 0.25 & 0.125 & 0.125 & 1e-4 & 1e-5 &  500 & 128 \\
cMNIST3D & 1 & 3 & -- & exp & 0.25 & 0.125 & 0.125 & 1e-4 & 1e-5 &  500 & 128 \\
cFashionMNIST & 1 & 10 & -- & exp & 0.25 & 0.125 & 0.125 & 1e-4  & 1e-5 & 500 & 128 \\
cChestMNIST & 1 & 4 & -- & exp & 0.1 & 0.02 & 0.02 & 1e-4 & 1e-5 &  500 & 128 \\ 
MNIST-LinearDecoder & 1 & 5 & -- & exp & -- & 0.1 & 0.1 & 5e-4 & 1e-5 &  2500 & 128 \\
\hline
\end{tabular}
}
\end{center}
\end{table*}
\clearpage

\clearpage
\putbib 

\end{bibunit}

\end{document}